\documentclass[10pt,twocolumn,letterpaper]{article}
\usepackage[pagenumbers]{cvpr}

\usepackage{times}
\usepackage{amsmath, amssymb}
\usepackage{graphicx}
\usepackage{hyperref}
\usepackage{colortbl}
\usepackage{multirow}
\usepackage{booktabs}
\usepackage{svg}
\usepackage{float}
\usepackage{multibib}
\newcites{appendix}{Appendix References}
\usepackage{color}
\usepackage{array}
\usepackage{tabularx}

\makeatletter
\newcommand\blfootnote[1]{
  \begingroup
  \renewcommand\thefootnote{}\footnote{#1}
  \addtocounter{footnote}{-1}
  \endgroup
}
\makeatother

\definecolor{linkcolor}{rgb}{0.21,0.49,0.74}
\hypersetup{
    colorlinks=true,
    linkcolor=linkcolor,
    citecolor=linkcolor,
    urlcolor=linkcolor
}

\title{V2PE: Improving Multimodal Long-Context Capability of \\ Vision-Language Models with Variable Visual Position Encoding}

\author{
    \textbf{Junqi Ge}\textsuperscript{1,4$*$},
    \textbf{Ziyi Chen}\textsuperscript{1,4$*$},
    \textbf{Jintao Lin}\textsuperscript{3,4$*$},
    \textbf{Jinguo Zhu}\textsuperscript{4$*$},\\
    \textbf{Xihui Liu}\textsuperscript{3},
    \textbf{Jifeng Dai}\textsuperscript{1,4},
    \textbf{Xizhou Zhu}\textsuperscript{1,2,4\dag}\\
    [2mm] 
    \textsuperscript{1}Tsinghua University \quad
    \textsuperscript{2}SenseTime Research \quad
    \textsuperscript{3}University of Hong Kong \\
    \textsuperscript{4}Shanghai AI Laboratory\\
    [2mm]
}

\begin{document}

\maketitle

\blfootnote{\noindent$^{*}$Equal contribution, in random order. \textsuperscript{\dag}Corresponding author: Xizhou Zhu \textless zhuwalter@sensetime.com\textgreater.}

\begin{abstract}

Vision-Language Models (VLMs) have shown promising capabilities in handling various multimodal tasks, yet they struggle in long-context scenarios, particularly in tasks involving videos, high-resolution images, or lengthy image-text documents. 
In our work, we first conduct an empirical analysis of the long-context  capabilities of VLMs using our augmented long-context multimodal datasets. 
Our findings reveal that directly applying the positional encoding mechanism used for textual tokens to visual tokens is suboptimal, and VLM performance degrades sharply when the position encoding exceeds the model's context window.
To address this, we propose Variable Visual Position Encoding (V2PE), a novel positional encoding approach that employs variable and smaller increments for visual tokens, enabling more efficient management of long multimodal sequences.
Our experiments demonstrate the effectiveness of V2PE to enhances VLMs' ability to effectively understand and reason over long multimodal contexts.
We further integrate V2PE with our augmented long-context multimodal datasets to fine-tune the open-source VLM, InternVL2. The fine-tuned model achieves strong performance on both standard and long-context multimodal tasks. Notably, when the sequence length of the training dataset is increased to 256K tokens, the model is capable of processing multimodal sequences up to 1M tokens, highlighting its potential for real-world long-context applications.
The code and models will be available at \href{https://github.com/OpenGVLab/V2PE}{https://github.com/OpenGVLab/V2PE}.
\end{abstract}

\section{Introduction}
\label{sec:intro}

\begin{figure}[htbp]
    \centering
    \includegraphics[width=0.88\linewidth]{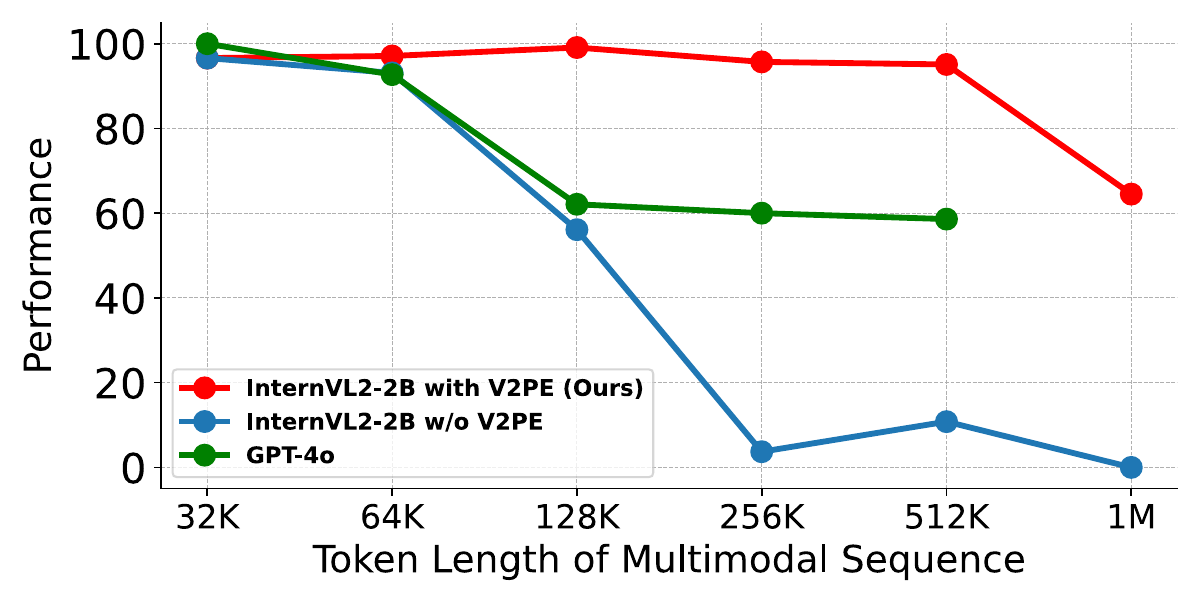}
\caption{Performance on the image retrieval task using a token length of up to 1M on MM-NIAH~\cite{mmniah} across different VLMs. The GPT-4o~\cite{gpt4o_web} version is 2024-08-06, while InternVL2-2B~\cite{internvl_1_5} models are both fine-tuned from the official release, using our augmented data, with token lengths reaching up to 256K per sample.}
    \label{fig:1mtoken}
\end{figure}

With the rapid advancement of Large Language Models (LLMs)~\cite{GPT4_report, llama2,internlm,moss,internlm2_report}, Vision-Language Models (VLMs) have made substantial strides~\cite{gpt4v,gemini_1_5,internvl,Qwen2VL}, excelling at tasks like visual captioning~\cite{chen2015microsoft}, visual question answering~\cite{DocVQA}, and complex visual reasoning~\cite{CLEVRER}. Despite this progress, existing research~\cite{mmniah,mmneedle,LongVA,LongVILA} reveals that VLMs struggle to generalize effectively when confronted with multimodal long-sequence inputs (e.g., long videos~\cite{Longvlm,Videoagent}, high-resolution images~\cite{Cogagent,Vary}, and lengthy image-text documents~\cite{Mm-interleaved,Flamingo,mmc4}). This limitation emerges even in relatively straightforward, such as object counting and passkey duplication, significantly restricting VLMs' potential applications and impeding the enhancement of user experience~\cite{LongLLaVA,mmniah,mobility_vila,wang2024enhancing}.

Recent efforts have attempted to extend VLMs' capabilities to process multiple images or handle long multimodal sequences. However, these approaches either permit only a small number of images (typically fewer than five)~\cite{mantis,llavanext,minigpt} or primarily target video data (e.g., LongVA~\cite{LongVA}, LongVILA~\cite{LongVILA}, LongLLAVA~\cite{LongLLaVA}).
These studies are only limited to specific application scenarios, highlighting the challenges VLMs face in handling complex and long-sequence multimodal data, which makes a key research question particularly urgent: \textit{Why do VLMs perform poorly in long-context scenarios, and how can we unlock their capacity for comprehensive multimodal understanding and reasoning over long sequences?}

To investigate this, we first construct a large-scale pool of long-context multimodal datasets to evaluate and analyze VLM capabilities systematically.
By extending the sequence length of existing instruction-tuning datasets (e.g., DocVQA~\cite{DocVQA}, ChartQA~\cite{ChartQA}, SQA~\cite{ScienceQA}) to 32K or 256K tokens, we adapt these datasets specifically to train VLMs for enhanced long-context capabilities. We also expand the validation sets to 64K and 1M tokens to validate VLM performance with longer contexts, providing a deeper understanding of the limitations. 

\begin{figure}
    \centering
    \includegraphics[width=0.9\linewidth]{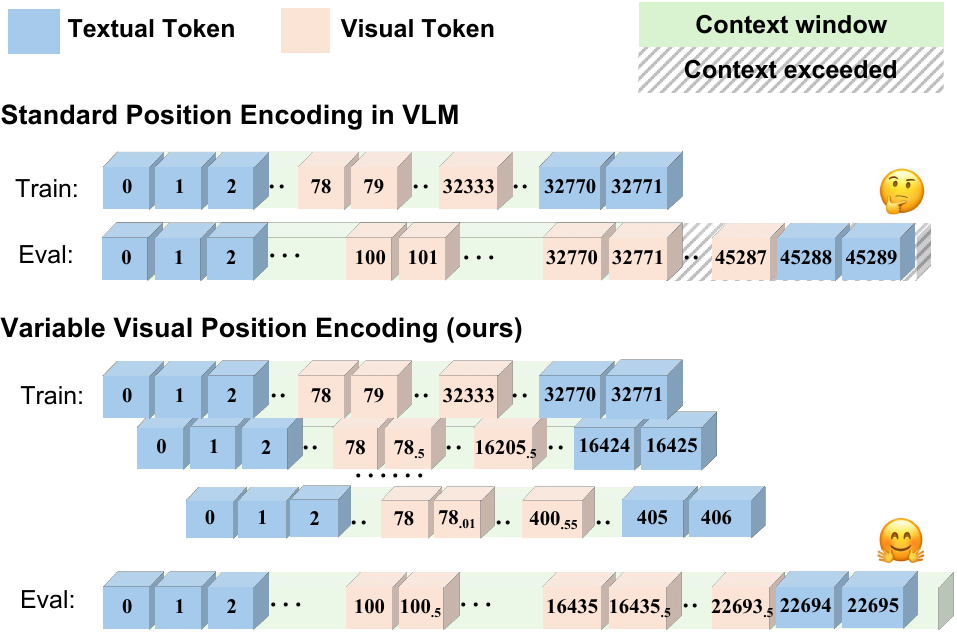}
    \caption{Illustration from our proposed Variable Visual Position Encoding (V2PE). Unlike the standard position encoding used in most VLMs, which shares the same stepwise positional increment for both visual and textual tokens, our proposed Variable Visual Positional Encoding (V2PE) uses smaller and variable positional increments specifically for visual tokens compared to textual tokens. This flexible design enables VLMs to handle long-context multimodal sequences within a limited context window. }
    \label{fig:enter-label}
\end{figure}

Our empirical analysis shows that the design of positional encodings of visual tokens plays an essential role in long-context scenarios, which is often overlooked in previous studies.
Specifically, (1) directly applying the LLM positional encoding mechanism to visual tokens is suboptimal, and (2) the performance of VLMs degrades significantly when positional encodings for visual tokens exceed the trained context window, and previous position encoding extension methods applicable to LLMs provide only marginal improvements.
These findings highlight the need for specialized positional encoding methods to manage visual tokens in long-context multimodal scenarios effectively.

To address these challenges, we propose Variable Visual Position Encoding (V2PE), a novel approach for handling visual token positions in VLMs. Considering the continuity nature of pixel space, adjacent visual tokens exhibit greater similarity compared to adjacent text tokens. Thus, V2PE uses smaller positional increments for visual tokens than for text tokens (see Fig.~\ref{fig:enter-label}). Furthermore, during training, V2PE employs variable positional increments for visual tokens, enabling the model to learn and adapt to position encoding in various scenarios. This variable adjustment allows the model to effectively handle different numbers and complexities of image inputs during inference, thereby enhancing its stability and adaptability in long-context processing.

In experiments, we apply V2PE to enhance the long-context capability of an open-source high-performance VLM, InternVL2-2B~\cite{internvl,internvl_1_5}, and fine-tune it using our extended multimodal datasets. The resulting model not only maintains strong performance on standard short-context multimodal benchmarks, but also excels in tasks requiring long context handling, which outperforms traditional token compression and other position encoding extension methods.  In particular, after further fine-tuning on multimodal sequences up to 256K tokens, our model achieves promising performance in multimodal retrieval tasks involving sequences as long as 1M tokens, as shown in Fig.~\ref{fig:1mtoken}.
The main contributions of this paper are as follows:
\begin{itemize}
\item  We construct mixed datasets for VLMs' long-context training and evaluation by augmenting existing multimodal instruction tuning datasets and conduct a thorough investigation into why current VLMs struggle with long-context multimodal inputs, revealing that directly applying LLM positional encoding to visual tokens is ineffective. 
\item   We propose Variable Visual Position Encoding (V2PE), a novel positional encoding strategy that employs variable and smaller increments for visual tokens, significantly enhancing VLMs' ability to understand and reason over long multimodal contexts.
\item  We apply our V2PE method and extend training data on the open-source VLM, InternVL2-2B. The fine-tuned VLM performs exceptionally well on both general multimodal benchmarks and long-context multimodal tasks, with the capacity to handle sequences of up to 1M tokens.
\end{itemize}

\section{Related Works}
\label{sec:related}

\noindent\textbf{Vision-Language Models (VLMs).}
With the development of Large Language Models (LLMs)~\cite{GPT4_report,Qwen_report,baichuan2,Deepseek,internlm2_report,Glm,moss,alpaca,internlm,LLaMA,llama2,cot,Skywork,zheng2023judging}, the integration of vision and language modalities is significantly catalyzed. These progress give rise to Vision-Language Models (VLM), which can perceive visual contents, conduct visual reasoning, and engage in multi-modal dialogue with humans. Both proprietary commercial VLMs~\cite{claude3series,HyperGAI,Mm1,Step-1V,grokv,gpt4v,gemini,gemini_1_5} and open-source VLMs~\cite{hu2024mplug,luo2024feast,tong2024eyes,Kosmos-2.5,wang2023see} has witnessed to this significant evolution. For commercial entities, GPT-4V~\cite{gpt4v} incorporates visual inputs to extend GPT-4~\cite{GPT4_report} with capabilities of handing multi-modal content, while Google’s Gemini series~\cite{gemini,gemini_1_5} are able to process 1 million multi-modal tokens with significant performance. For open-source initiatives, BLIP series~\cite{Blip,blip2,Instructblip}, LLaVA series~\cite{llavanext,improvedllava,llava}, Qwen-VL series~\cite{Qwen-VL,Qwen2VL}, MiniCPM-V series~\cite{MiniCPM-V}, InternVL series~\cite{internvl,internvl_1_5,mini_internvl},
and others~\cite{KOSMOS,Kosmos-2,all-seeing,internlmxcomposer2_5,internlmxcomposer2_4khd,internlmxcomposer2,internlmxcomposer,mobilevlm,mobilevlm2,Mini-gemini,deepseek-vl,Paligemma} have also impacted the AGI landscape in the research community by linking large language models~\cite{LLaMA,internlm2_report,Qwen_report} and large vision models~\cite{CLIP,vit,dehghani2023scaling,internvl} in processing both visual and textual modality.
Specifically, there are also some works to advance VLMs' power by improving scale and quality of data~\cite{Infinity-MM,sharegpt4v}, training on high-resolution images~\cite{Cogagent,Vary}, optimizing vision foundation models~\cite{internvl,sigLIP}.

\label{sec:method}
\begin{figure}[t]
    \centering
    \begin{minipage}{0.47\textwidth}
        \centering
        \includegraphics[width=\textwidth]{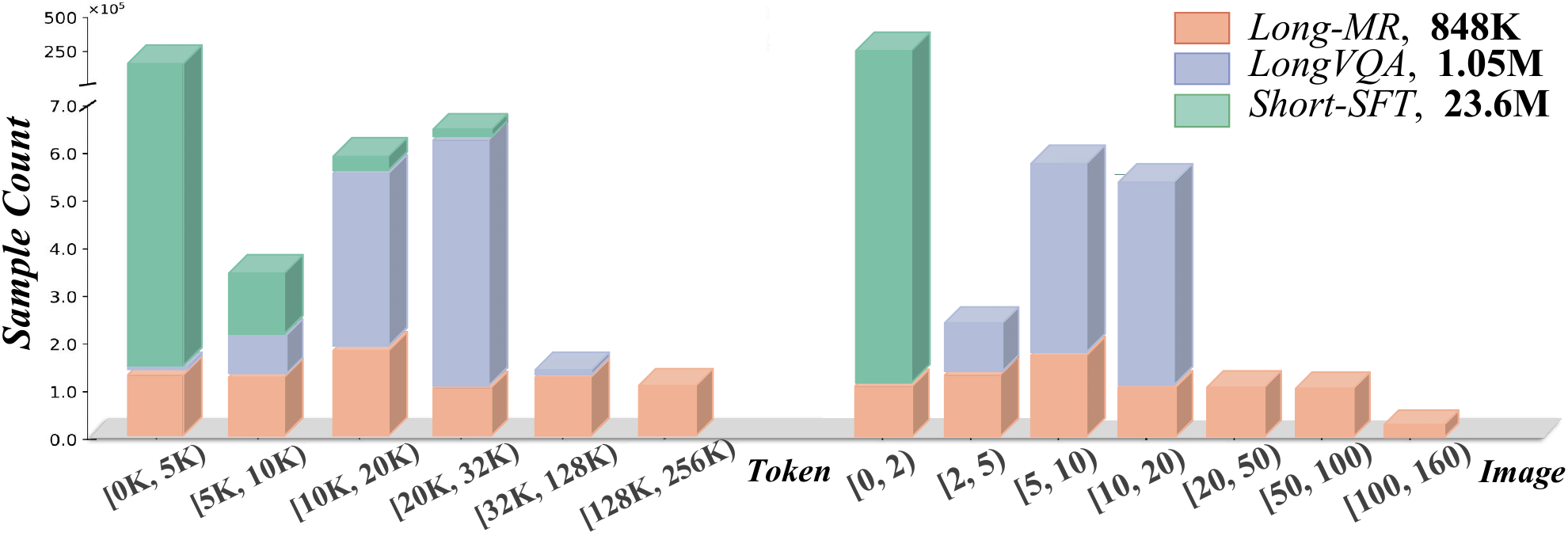}
    \end{minipage}
    \caption{Statistics of our mixed training dataset, including Long-MR, Long-VQA, and short SFT data from InternVL2. The left part and the right part illustrate the distribution of tokens per sample and images per sample, respectively.}
    \label{fig:count2}
\end{figure}
\vspace{0.5em}

\noindent\textbf{Long Context Modeling.}
Due to practical demands, a large body of research~\cite{soaring,LongSkywork,E2-LLM,Longrecipe,rmt,chevalier2023adapting} has been developed to extend the context length of LLMs. One mainstream direction is to increase the length of LLM's context window by extrapolation of position encoding~\cite{ntk,ALiBi,PI,ReRoPE,ReFormer_pe,YaRN,LongroPE,zhang2024extending,PoSE}, which allows the LLM to process unseen position during inference. Another line of research focuses on alleviating complexity of self attention by using sparse attention~\cite{LongloRA,child2019generating,Longformer,BigBird,Longnet,blockwise}, linear-complexity state-space modules~\cite{gu2021efficiently,mamba}, or approximate attention computation~\cite{Reformer,Linformer,performers,Combiner} as the alternatives for dense global attention. The context can also be extended by utilizing external memory to retrieve relevant tokens from the compression of past inputs, such as chunked input~\cite{landmark_attn,aug_llm_long,xu2023retrieval,zhang2023retrieve} or the cached KV activations~\cite{memorizing,focused_transformer,selective_cache,Keyformer,willette2024training}. Other works~\cite{StreamingLLM,Lm-infinite,willette2024training} also adopt sliding windows to achieve an infinite context by only maintaining the activations for the very first and the latest tokens.
Recent efforts~\cite{LongVA,LongVILA,LongLLaVA,mantis} aim to improve VLMs' ability to process multiple images and long multimodal sequences. However, these studies are often limited to specific applications, revealing challenges faced by VLMs in processing  complex and long-context data.

\vspace{0.5em}
\noindent\textbf{Position Encoding in Transformer.}
To address the lack of sequential information inherent in self-attention mechanisms, Transformers~\cite{transformer} utilize position encoding as a fundamental component to provide position information. The common position encoding design can be broadly categorized into absolute position encoding (APE) and relative position encoding (RPE). The absolute position encoding~\cite{transformer,wang2020position,kiyono2021shape,GPT-3,Opt,ke2020rethinking} is simple and intuitive, as it embeds each absolute position position into a position vector which is then added into the representation of input sequences, but this strategy cannot be applied effectively for long sequences. To improve the modeling of long-term dependencies, relative positional encoding~\cite{shaw2018self,raffel2020exploring,lv2023we,mt5,Transformer-xl,ALiBi,sun2022length,rope} focuses on utilizing distance between tokens as the position information. Some research works~\cite{ALiBi,sun2022length,haviv2022transformer} aim to utilize relative position for length extrapolation to train Transformers on short sequences and inference on longer context. 
Another research direction is Rotary Position Encoding (RoPE), which interpolates position information by rotating the key-query products according to their relative distances. 
Methods like Position Interpolation~\cite{PI}, NTK-Aware scaling~\cite{ntk}, and LongRoPE~\cite{LongroPE} make progress on top of RoPE~\cite{rope} to improve its poor extrapolation capability for longer sequences.
For VLMs, most studies~\cite{internvl_1_5,Qwen-VL,LongVILA} adopt the same position embedding methods used in LLMs, and there are also some works~\cite{internvl_1_5,Qwen-VL,LongVILA} to explore different diagrams of the positional encoding schemes.

\section{Method}

\subsection{Augmented Long-context Multimodal Datasets}
\label{sec:dataset}
We introduce two augmented long-context multimodal datasets: \textit{Long Visual Question Answering}  and \textit{Long multimodal Retrieval}. These datasets aim to enhance VLMs' long-context training and establish a systematic evaluation framework,  thereby addressing the challenges associated with long-context understanding that extend beyond the scope of existing training data.

\vspace{0.2em}

\noindent \textbf{Long Visual Question Answering (Long-VQA).} The Long-VQA dataset aims to evaluate the capabilities of VLMs in understanding and reasoning over long multimodal sequences within general visual question-answering tasks. We extended 17 widely adopted datasets (e.g., DocVQA~\cite{DocVQA}, GQA~\cite{GQA}, SQA~\cite{ScienceQA}), expanding their content from short sequences to those containing up to 32K tokens. The tasks involve answering questions that require commonsense reasoning, factual knowledge, and interpretation of visual information from charts, documents, and real-world texts.

To extend existing datasets, we interleaved images from multiple samples into a single input, increasing sequence length and complexity to simulate real-world scenarios involving extraneous information. 
To reduce ambiguity, we refined questions to be more specific, incorporating instructions like ``Based on image $n$, answer the following question'', which helps models focus on relevant content.

Long-VQA contains 533K samples: 392K for training (up to 32K tokens) and 141K for validation (up to 64K tokens) to evaluate the generalization to longer contexts. 

\vspace{0.2em}
\noindent \textbf{Long Multimodal Retrieval (Long-MR).} 
Inspired by MM-NIAH (Multimodal Needle-in-a-Haystack)~\cite{mmniah}, we developed Long-MR by inserting a target image or textual segment into sequences of interleaved images and texts. Long-MR evaluates VLMs' ability to retrieve specific targets from ultra-long multimodal sequences, requiring models to locate the inserted ``needle'' and answer associated questions.            
We generated two subsets of Long-MR: Long-MR-32K (488K samples, sequences up to 32K tokens) and Long-MR-256K (50K samples, sequences up to 256K tokens), following the data construction process of MM-NIAH.

To assess the limits of VLMs' long-context capabilities, we further extend the official MM-NIAH evaluation benchmark by generating testing samples with sequence lengths ranging from 64K to 1M tokens, resulting in the MM-NIAH\textsubscript{1M} benchmark. This extension pushes the testing capacity beyond the original MM-NIAH, which was limited to sequences of up to 64K tokens.

We combine the training splits of Long-VQA and Long-MR with short-context instruction-tuning datasets, as utilized in InternVL2, to create a mixed training set for our experiments. Fig.~\ref{fig:count2} illustrates the distribution of token sequence lengths and the number of images in the mixed training dataset, highlighting the significant increase in long-context samples introduced by our augmented datasets.

For more details about the construction process, data examples, and statistics of these datasets, please refer to the appendix.

\subsection{Variable Visual Position Encoding}

\paragraph{Position Encoding in Vision-Language Models.}

Position encoding is essential in Transformer architectures, enabling models to capture sequential relationships by providing tokens with positional information. It usually involves two sequential steps: \textit{Position Index Derivation} \( f_{\text{pos}} \), which assigns a positional index \( p_i \) to each token \( x_i \), and \textit{Position Embedding Computation} \( g_{\text{emb}} \), which transforms these indices into position embeddings that influence the attention mechanism.

Formally, the multimodal input in VLMs can be represented as a sequence of \( N \) interleaved textual and visual tokens:
\begin{equation}
\mathbf{X} = [x_0, x_1, \dots, x_{N-1}],
\end{equation}
where each token \( x_i \) is either a textual token \( x_i^{\text{txt}} \) or a visual token \( x_i^{\text{vis}} \).

The Position Index Derivation function \( f_{\text{pos}} \) is recursively defined to capture the sequential nature of token positions:
\begin{equation}
p_i = \begin{cases}
0, & \text{if } i = 0, \\
f_{\text{pos}}(p_{i-1}, x_i), & \text{for } i = 1, 2, \dots, N-1.
\end{cases}
\label{eq2}
\end{equation}
In existing LLMs and VLMs, the position index increments uniformly by 1 for each token, regardless of its modality:
\begin{equation}
p_i = p_{i-1} + 1, \quad \text{for } i = 1, 2, \dots, N-1.
\label{eq3}
\end{equation}

The Position Embedding Computation \( g_{\text{emb}} \) then transforms these position indices into embeddings. VLMs typically adopt the same position embedding methods used in Large Language Models (LLMs), such as Relative Position Encoding~\cite{shaw2018self} or Rotary Position Embedding (RoPE)~\cite{rope}. These embeddings are incorporated into the token representations to provide positional context during attention computations. For instance, in RoPE encoding, the token representation \( \mathbf{h}_i \) integrates positional information as:
\begin{equation}
\mathbf{h}^{'}_i = \mathbf{h}_i \otimes g_{\text{emb}}(p_i),
\end{equation} 
where  $\otimes$ represents element-wise multiplication, and  \( \mathbf{h}^{'}_i \) is subsequently used in the Transformer attention mechanism.

\begin{figure*}[h]
    \begin{minipage}{0.31\textwidth}
        \centering
        \includegraphics[width=\linewidth]{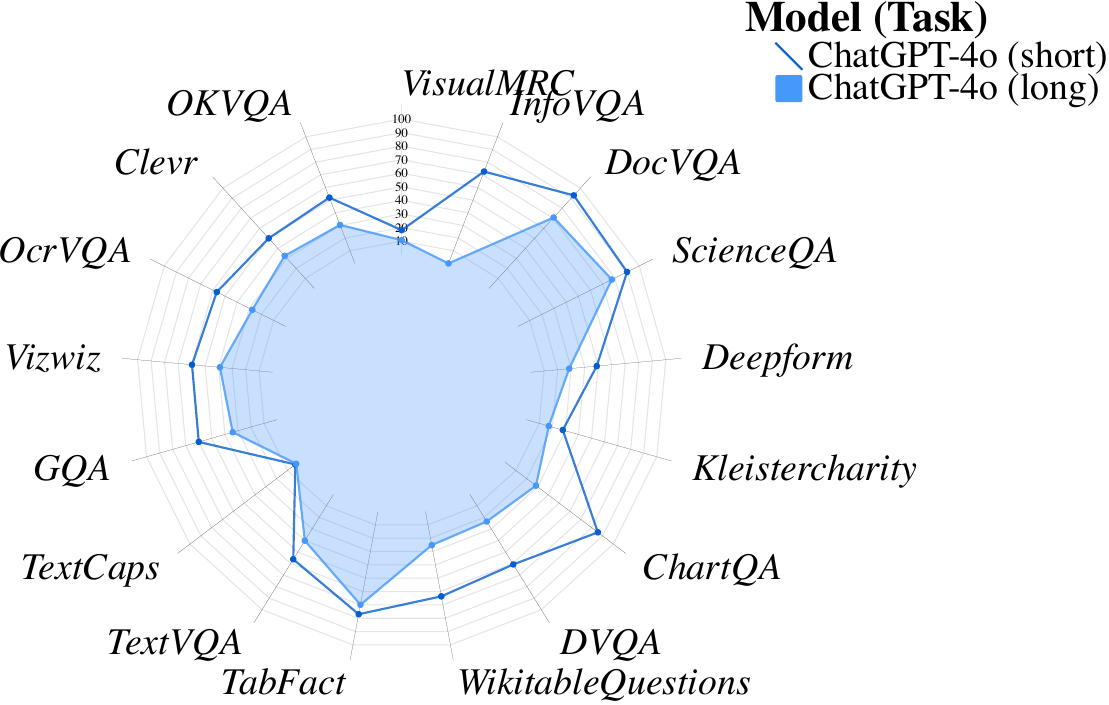}
    \end{minipage}
    \hfill
    \begin{minipage}{0.33\textwidth}
        \centering
        \includegraphics[width=\linewidth]{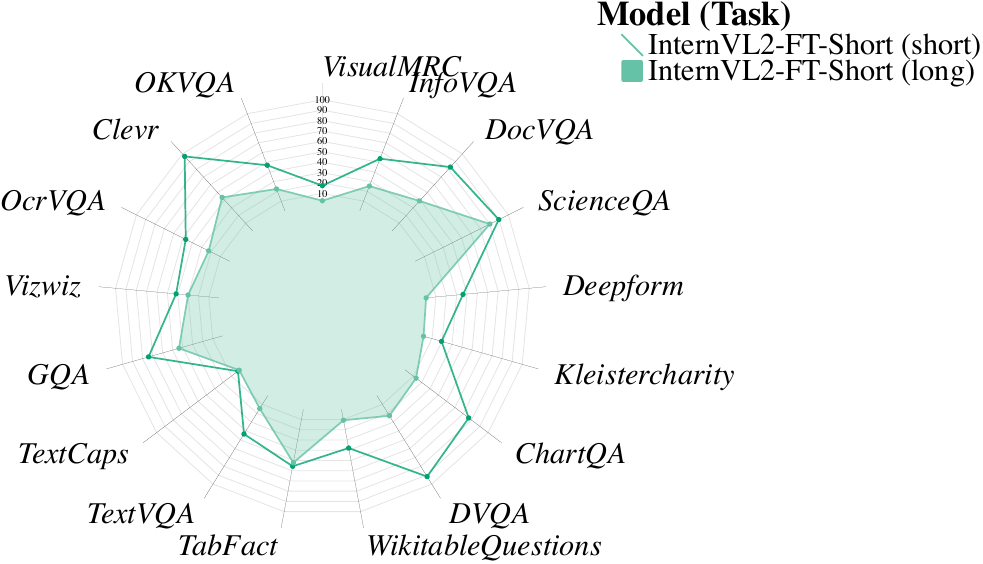}
    \end{minipage}
     \hfill
    \centering
    \begin{minipage}{0.32\textwidth}
        \centering
        \includegraphics[width=\linewidth]{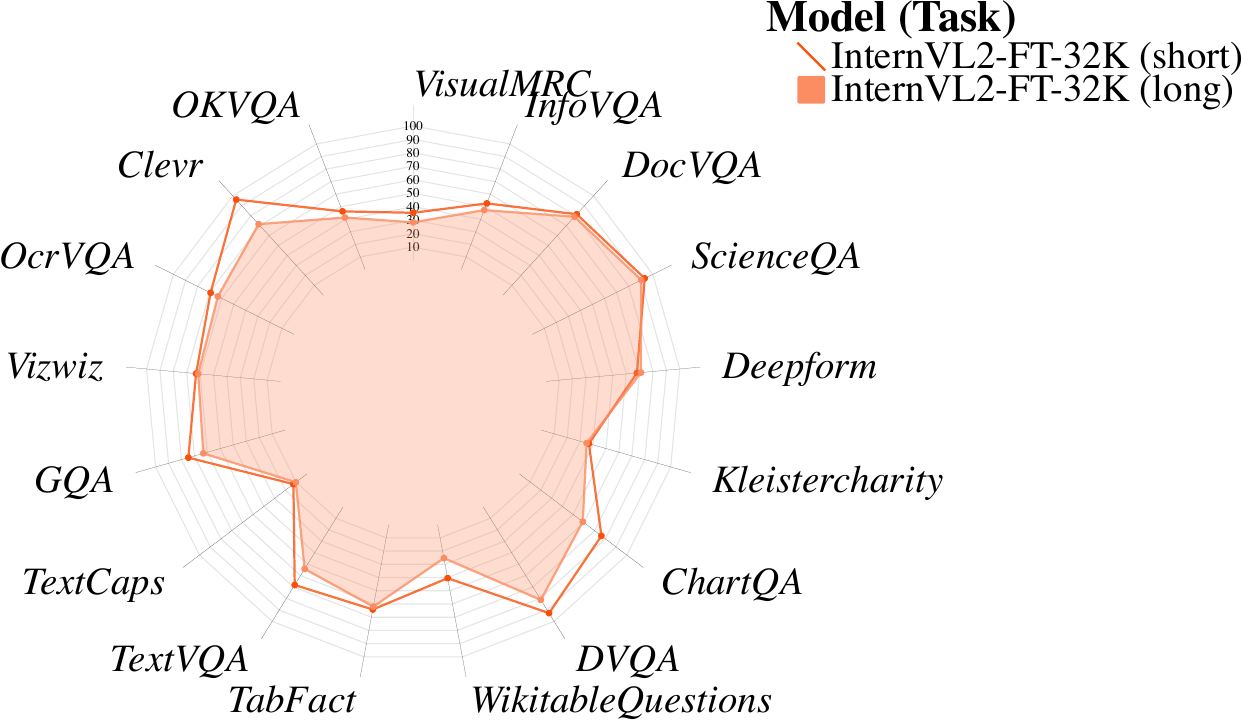}
    \end{minipage}
   
    \caption{Performance on Long-VQA (long) with sequence lengths up to 32K tokens and its corresponding standard VQA (short) benchmarks. Compared with GPT-4o and InternVL2-FT-Short, the InternVL2-FT-32K, enhanced with our proposed V2PE, demonstrates comparable outstanding performance across both standard VQA benchmarks and their long-context counterparts. }
    \label{fig:radar}
\end{figure*}

\paragraph{Variable Position Index Derivation.}

The uniform increment of position indices in current VLMs does not account for the differences in information complexity and redundancy between textual and visual tokens. Visual tokens often exhibit higher redundancy and greater similarity with adjacent tokens, suggesting they may require smaller positional increments than textual tokens. Moreover, the large number of visual tokens can cause position indices to exceed the model's pre-trained context window, leading to degraded performance.

To address these issues, we propose a modality-specific recursive function for position index derivation, assigning position indices differently for textual and visual tokens:
\begin{equation}
p_i = p_{i-1} + \begin{cases}
1, & \text{if } x_i \text{ is a textual token}, \\
\delta, & \text{if } x_i \text{ is a visual token},
\end{cases}
\label{eq:delta}
\end{equation}
where \( \delta \) is a smaller increment (\( \delta < 1 \)) that reduces the rate at which position indices increase for visual tokens. The standard increment of 1 is retained for textual tokens to maintain their positional distinctions.

During training in our experiments, \( \delta \) is dynamically selected for each image from a set of fractional values:
\begin{equation}
\delta \in \Delta = \left\{1, \frac{1}{2}, \frac{1}{4}, \frac{1}{8}, \frac{1}{16}, \frac{1}{32}, \frac{1}{64}, \frac{1}{128}, \frac{1}{256}\right\}.
\label{delta_set}
\end{equation}
Note that, \( \delta \) remains constant within a single image to preserve the relative positional relationships among its visual tokens. For inputs containing multiple images, \( \delta \) can be independently chosen for each image.

During inference, \( \delta \) can be flexibly selected based on the input sequence length, allowing us to balance task performance and ensure that position indices remain within the model's valid context range. Specifically, for long input sequences, a smaller \( \delta \) can be employed to control the increase in position indices, preventing them from exceeding the trained positional embedding range, as illustrated in Fig.~\ref{fig:enter-label}.

\paragraph{Discussion and Comparison with Previous Methods.}

Our Variable Visual Position Encoding (V2PE) offers several advantages over existing long-context methods:

\noindent 1)  Unlike approaches that reduce the number of visual tokens through attention pooling or feature pooling—which potentially lead to information loss—V2PE retains all visual tokens within the VLM, preserving the richness and granularity of visual content.

\noindent 2) Position encoding extension methods (\emph{e.g.,} Positional Interpolation)  adjust position embeddings during inference to accommodate longer sequences. However, this can introduce inaccuracies due to extrapolation beyond the trained positional embedding range. In contrast, V2PE allows the VLM to adapt position indices with arbitrary intervals by dynamically selecting \( \delta \) during training. This strategy avoids unexpected position embeddings and ensures consistent model performance across varying input lengths.

\begin{figure}[t]
    \begin{minipage}{0.23\textwidth}
        \centering
        \includegraphics[width=\linewidth]{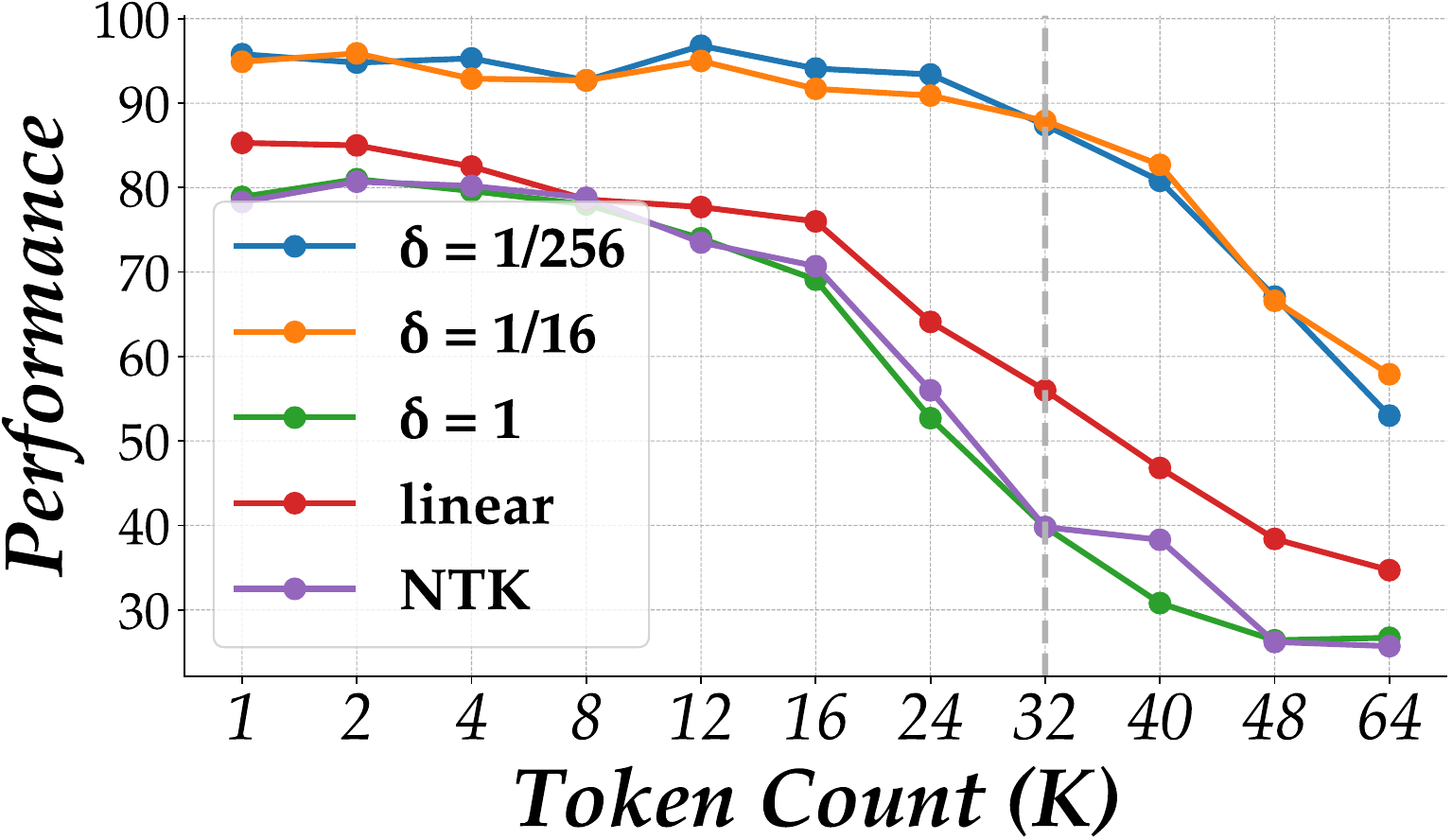}
    \end{minipage}
    \hfill
    \centering
    \begin{minipage}{0.23\textwidth}
        \centering
        \includegraphics[width=\linewidth]{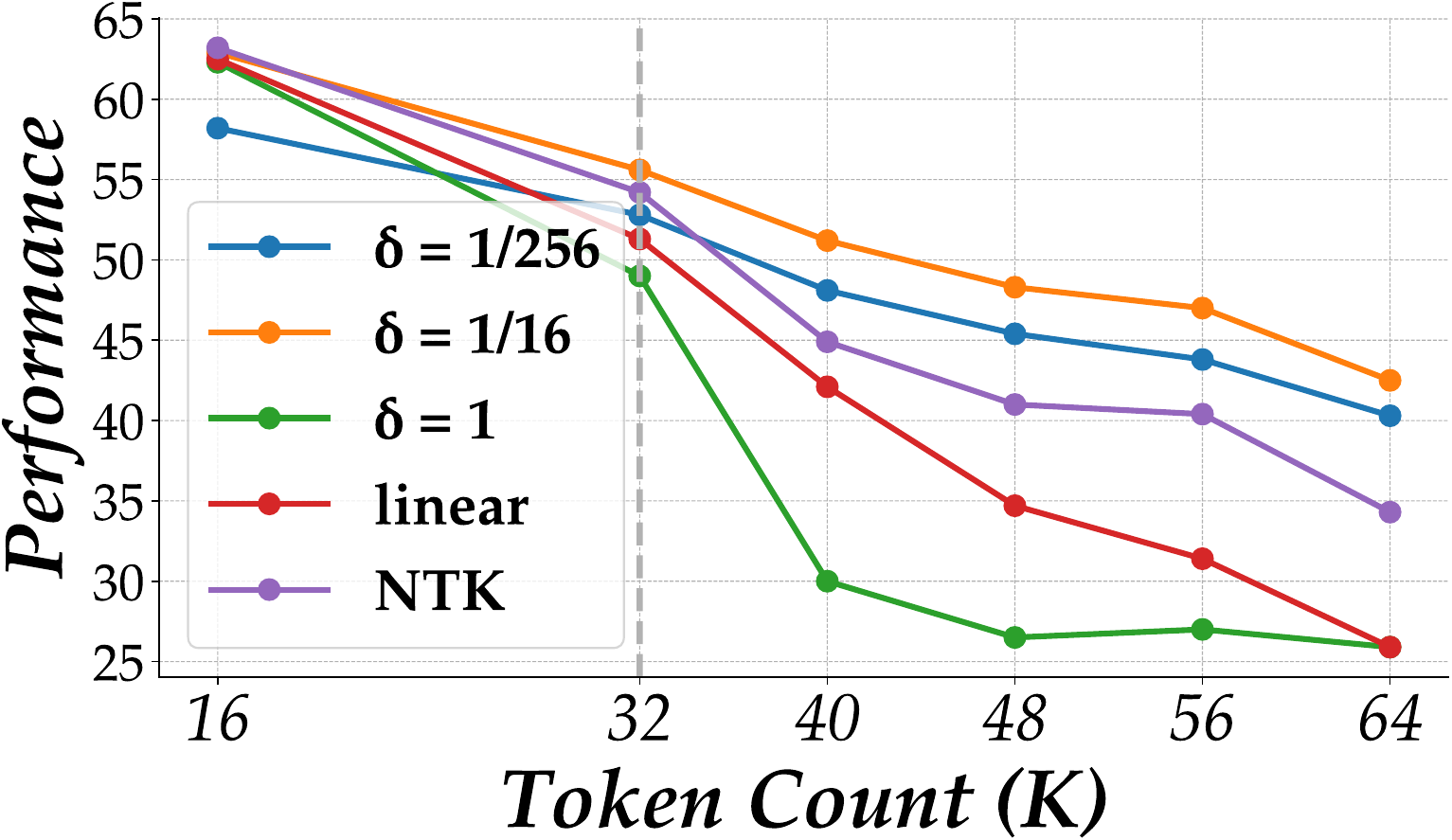}
    \end{minipage}
       \vspace{-0.6em}
    \caption{Performance on image retrieval task in MM-NIAH (left) and QA task in Long-VQA (right) with different positional increments.  Additionally, we include the results of InternVL2-FT-32K when utilizing  linear interpolation  (linear)~\cite{PI} and   NTK-Aware Scaled RoPE (NTK)~\cite{ntk} position encoding extension methods. }
    \label{fig:stride_line}
\end{figure}

\begin{table}[t]
    \centering
    \caption{
    Performance on Long-VQA task with sequence lengths up to 64K tokens at various  positional increments  $\delta$.}
        \vspace{-0.6em}
    \resizebox{0.97\linewidth}{!}{
    \begin{tabular}{l c|c c c c c c | c } 
    \toprule
        Model   & $ \delta$ & 16K & 32K & 40K & 48K & 56K & 64K & Avg \\ \midrule 
    \multirow{1}{*}{ InternVL2-2B }
        & $-$   & 52.8 & 27.3 & 23.0 & 23.9 & 22.6 & 21.0 &28.4 \\ 
    \midrule
    \multirow{9}{*}{InternVL2-V2PE-32K} 
        & 1/256   & 58.2 & 52.8 & 48.1 & 45.4 & 43.8 & 40.3 & 48.1 \\
        & 1/128  & 60.0 & 53.7 & 49.5 & 46.5 & 45.3 & 41.3 & 49.4 \\
        & 1/64  & 61.0 & 54.5 & 50.1 & 47.4 & 45.9 & 42.2 & 50.2 \\
        & 1/32   & 61.9 & 55.4 & 51.4 & 47.5 & 46.4 & 42.3 & 50.8 \\
        & 1/16  & 62.9 & 55.6 & 51.2 & \textbf{48.3} & \textbf{47.0} & \textbf{42.5} & \textbf{51.3} \\
        & 1/8 & 63.2 & \textbf{55.8} & \textbf{52.0} & 47.6 & 45.8 & 41.9 & 51.1 \\
        & 1/4 & \textbf{63.4} & 55.7 & 51.3 & 47.0 & 45.1 & 39.7 & 50.4 \\
        & 1/2 & 63.3 & 54.9 & 46.8 & 36.8 & 33.3 & 29.2 &44.1 \\
        & 1/1 & 62.3 & 49.0 & 30.0 & 26.5 & 27.0 & 25.9 & 36.8 \\
    \midrule
    InternVL2-FT-32K (Linear~\cite{PI}) 
        & $-$   & 62.5 & 51.3 & 42.1 & 34.7 & 31.4 & 25.9 & 40.9 \\
    InternVL2-FT-32K (NTK~\cite{ntk}) 
        & $-$  & 63.2 & 54.2 & 44.9 & 41.0 & 40.4 & 34.3 & 45.3 \\
    
    \bottomrule
    \end{tabular}\label{tab:exp_longvqa}
    }

\end{table}

\begin{table*}[h]
    \centering

    \caption{Performance  on standard VQA benchmarks with different positional increment $\delta$. The highest score is marked in \textbf{bold}.}
        \vspace{-1.0em}
    \resizebox{0.9\linewidth}{!}{
    \begin{tabular}{c c| c c c c c c c c c | c} \toprule
Model   & $\delta$ & ChartQA & DocVQA & AI2D & InfoQA & SQA & POPE & MMMU\textsubscript{val} & MMBench\textsubscript{EN} & SEED\textsubscript{I} & Avg \\ \midrule 
    \multirow{1}{*}{ InternVL2-FT-Short }
        & $-$   & 76.0 & \textbf{85.7} & \textbf{74.0} & \textbf{57.1} & 94.0& \textbf{88.9} & 34.8 & 73.0 & 66.6 & 72.2 \\ \hline
    \multirow{9}{*}{InternVL2-V2PE-32K}
        & 1/256   & 76.2 & 81.6 & 72.3 & 51.7 & 94.3 & 88.0 & 35.7 & 73.2 & 71.0 & 71.6 \\
        & 1/128   & 76.5 & 82.3 & 72.6 & 52.9 & 94.2 & 88.1 & 34.7 & 73.5 & 70.9 & 71.7  \\
        & 1/64   & 76.4 & 83.0 & 72.6 & 53.6 & 94.4 & 88.0 & 35.4 & 73.4  & 70.8 & 72.0 \\
        & 1/32  & 76.9 & 83.6 & 72.6 & 54.3 & 94.3 & 88.1 & 35.7 & 73.3 & 71.0 &  72.2 \\
        & 1/16  & 77.2 & 84.3 & 73.0 & 55.2 & 94.5 & 88.1 & 35.7 & 73.3 & 71.0 & 72.5 \\
        & 1/8  & 77.0 & 84.6 & 73.1 & 55.4 & 94.6 & 88.1 & 36.0 & 72.9 &  71.0 & 72.5 \\
        & 1/4  & \textbf{77.3} & 84.7 & 73.3 & 56.5 & 94.6 & 88.2 & 36.0 & 73.6 & 71.0 & 72.8 \\
        & 1/2 & 76.8 & 84.9 & 73.5 & 56.0 & \textbf{94.9} & 88.7 & 35.8 & \textbf{73.8} & \textbf{71.2} & \textbf{72.9} \\
        & 1/1 & 76.6 & 84.9 & 73.7 & 55.2 & 94.8 & 88.4 & \textbf{36.1} & 73.7 & 71.1 & 72.7 \\ 
        \bottomrule
    \end{tabular}}\label{tab:v33_detail}
    \vspace{-0.2em}
\end{table*}

\begin{table*}[htbp]
    \centering
    \caption{Performance on the image retrieval task in MM-NIAH with token lengths up to 1M at various position increments $\delta$. Test samples exceeding 64K tokens are from our extended MM-NIAH\textsubscript{1M}. The designation ``NA'' indicates that the GPT-4 results are not applicable due to the extremely long context length.}
        \vspace{-1.0em}
    \label{tab:exp_mmniah}
    \resizebox{0.72\linewidth}{!}{
    \begin{tabular}{c c| c c c c c c c c} \toprule
        Model   & $\delta$ & 1K& 16K & 32K & 64K & 128K & 256K & 512K & 1M\\ 
        \midrule 
        \multirow{1}{*}{ InternVL2-2B }    
        & $-$ & 23.9 & 33.7 & 28.7 & 26.0 &17.3&21.8&5.3&0.0 \\ 
        \multirow{1}{*}{ GPT-4o }    
        & $-$ & 100.0 & 90.4 & 100.0 & 92.7 & 62.1 & 60.0 & 58.6 & NA \\ 
        \midrule
    \multirow{5}{*}{     InternVL2-V2PE-32K}
        & 1/256 & \textbf{100.0} & \textbf{84.7} & \textbf{79.3} & \textbf{81.5} & \textbf{61.2} & \textbf{56.9} & \textbf{36.7} & \textbf{6.0} \\
        & 1/64 & 96.0 & 83.9&  78.1 & 79.5 & 43.6 & 39.3 & 23.4 & 3.4 \\
        & 1/16 &  96.0 &  82.4 &  73.1 & 65.4 & 57.6 & 41.3 & 19.8 & 0.0 \\
        & 1/4 &  96.0 &  84.2 &  76.2 & 57.6 & 48.9 & 21.5 & 2.8 & 0.0 \\
        & 1/1 &  83.0 &  61.2&  42.0 & 27.5 & 25.5 & 1.5 & 0.0 & 0.0 \\
    \midrule
    \multirow{5}{*}{ InternVL2-V2PE-256K }  
        & 1/256 &  \textbf{96.0} & \textbf{ 97.1} &  \textbf{96.6} &  \textbf{97.1} &  \textbf{99.1} &  \textbf{95.7} & \textbf{95.1} & \textbf{64.5} \\
        & 1/64 &  96.0 & 94.7&  96.6 &  97.1 &  97.1 &  94.1 & 94.1 & 62.8 \\
        & 1/16 &  96.0 &  94.7&  93.1 &  94.2 &  100 &  96.2 & 92.9 & 45.1 \\
        & 1/4 &  96.0 & 94.4 &  100.0 &  97.1 &  75.5 &  75.2 & 38.3 & 2.5 \\
        & 1/1 &  91.0 &  92.8 &  96.6 &  93.1 &  56.1 &  3.7 & 10.8 & 0.0 \\
    \midrule
    \multirow{1}{*}{InternVL2-FT-32K  with linear interpolation} & $-$ & 85.3 & 76.0 & 56.0 & 34.7 & 26.4 & 21.9 & 16.3 & 0.5 \\ 
    \multirow{1}{*}{InternVL2-FT-32K with NTK interpolation} & $-$ & 78.3 & 70.7 & 39.8 & 25.7 & 12.3  & 16.8  &14.8&0.0 \\   
    \bottomrule 
    \end{tabular}
    \label{tab:mmniah_result}
                    }
                    \vspace{-0.6em}
\end{table*}

\section{Experiment}

\subsection{Analysis of VLMs' Long-Context Capabilities}

We analyze the long-context capabilities of existing VLM by using our constructed long-context multimodal datasets (see Sec.~\ref{sec:dataset}). Note that this sub-section does not use V2PE.

\vspace{0.5em}
\noindent\textbf{Experimental Setup.} We fine-tune InternVL2-2B on two different training datasets to create four finetuned models:
1) \textit{InternVL2-FT-Short}: This model is fine-tuned on the original instruction-tuning dataset (short-context) used by InternVL2-2B. It serves as a baseline to evaluate performance without exposure to long-context data.
2) \textit{InternVL2-FT-32K}: 
This model is fine-tuned on a mixed training dataset introduced in Sec.~\ref{sec:dataset}, incorporating our augmented long-context training data to enhance its long-context capacity.
3) \textit{InternVL2-32K-s1/16}: To explore the effect of positional encoding increments on visual tokens, we reduce the positional increment to $1/16$ for visual tokens, while maintaining training on the same mixed dataset as InternVL2-FT-32K.
4) \textit{InternVL2-32K-s1/256}: Similarly, we further reduce the positional increments to $1/256$ to examine its impact. 
Notably, the positional increments for these models remain fixed during training, without using the V2PE method.

\noindent\textbf{Evaluation Benchmarks.}  All models are evaluated on the Long-VQA validation split and the MM-NIAH benchmark. Additionally, performance on the standard multimodal VQA benchmarks is assessed if needed.

\noindent\textbf{Effectiveness of Augmented Long-Context Multimodal Data.} 
We compare the performance of InternVL2-FT-Short and InternVL2-FT-32K on both the Long-VQA validation set and standard short-context benchmarks, as shown in Fig.~\ref{fig:radar}. The results reveal that InternVL2-FT-Short, which is trained exclusively on short-context data, performs competitively on standard benchmarks but suffers significant degradation on long-context tasks. Even advanced models, such as GPT-4o, exhibit notable declines when the input token count is increased, underscoring a common limitation in handling extended sequences.
In contrast, InternVL2-FT-32K, trained on the mixed dataset including long-context data, effectively narrows the performance gap between short and long-context tasks. This suggests that VLMs cannot inherently generalize to long-context multimodal tasks from short-context training alone; targeted exposure to long-context data is crucial for achieving robust performance across varying input lengths.

\noindent\textbf{Impact of Visual Position Encoding.} 
To assess the influence of position encoding increments for visual tokens,  we evaluate the models InternVL2-FT-32K, InternVL2-32K-s1/16, and InternVL2-32K-s1/256 on both the Long-VQA and MM-NIAH benchmarks.

Experimental results, presented in Fig.~\ref{fig:stride_line}, indicate that InternVL2-FT-32K continues to experience performance degradation as input token sequences grow 8K token, despite being trained on sequences up to 32K tokens. When the input sequence length exceeds the token counts seen during training, performance declines sharply, and even applying positional encoding extension techniques developed for large language models (LLMs) provides only limited benefits.
However, models with reduced positional increments for visual tokens show significant improvements. Both InternVL2-32K-s1/16 and InternVL2-32K-s1/256 maintain stable performance. 
We hypothesis that reducing the position increments by factors of 16 and 256 effectively prevents the position indices of visual tokens from exceeding the model's trained context window, thereby mitigating performance decline.

\begin{table*}[t!]
    \scriptsize
    \caption{Comparison with existing MLLMs on  general MLLM benchmarks. ``\#Param'' denotes the number of parameters. The designation ``$-$'' indicates that the corresponding score is not released. }
    \vspace{-1.0em}
    \centering
    \setlength\tabcolsep{2pt}
    \renewcommand{\arraystretch}{1.15}
    \resizebox{0.78\linewidth}{!}{
    \begin{tabular}{l c|c c c c c c c c c|c}
    \toprule
    Model         & \#Param            & ChartQA       & DocVQA        & AI2D          & InfoVQA       & SQA           & POPE          & MMMU\textsubscript{val}      & MMBench\textsubscript{EN}  & SEED\textsubscript{I}        & Avg \\
    \midrule
    InternVL2-2B~\cite{internvl_1_5}   & 2.0B                 & 71.7          & 86.9          & 74.1          & 58.9          & 94.1          & 85.2          & 36.3            & 73.4          & 70.9          & 72.4 \\
    DeepSeek-VL-1.3B~\cite{deepseek-vl}  & 2.0B               & 47.4          & $-$          & 51.5          & $-$          & 68.4          & 85.9          & 33.8            & 66.4          & 66.0          & $-$ \\
    Qwen2-VL-2B~\cite{Qwen2VL}    & 2.0B                  & 73.5          & 90.1          & 74.7          & 65.5          & $-$           & $-$           & 41.1            & 74.9          & $-$           & $-$ \\
    Aquila-VL-2B~\cite{Infinity-MM}    & 2.2B               & 32.0          & 85.0          & 75.1          & 58.3          & 95.1          & 83.1          & 46.9            & 79.0          & 73.9          & 69.8 \\
    MiniCPM-V-2~\cite{MiniCPM-V}     & 2.8B               & 55.6          & 71.9          & 62.9          & $-$          & 80.7          & 86.3          & 38.2            & 64.1          & 67.1          & $-$ \\
    Vintern-3B-beta~\cite{Vintern} & 3.7B               & 68.3          & $-$          & 69.1          & $-$          & 75.0          & 87.4          & 46.7            & 70.6          & 70.0          & $-$ \\
    Llama 3.2 11B~\cite{llama2}    &  11B              & 83.4          & 88.4          & 91.1          & $-$           & $-$           & $-$           & 50.7            & 68.0           & $-$           & $-$ \\
    Qwen2-VL-72B~\cite{Qwen2VL}      & 73B            & 88.3          & 96.5          & 88.1          & 84.5          & 91.2          & 87.2          & 64.5            & 86.9          & 77.9          & 85.0 \\
    GPT-4o~\cite{gpt4o_web}  & $-$ & 85.7          & 92.8          & 84.7          & $-$           & 90.1          & 97.2          & 69.1            & 82.1          & 76.7          & $-$ \\
    \rowcolor{gray!15}
    \textbf{InternVL2-V2PE-32K}          & 2.0B        & \textbf{77.3} & \textbf{84.9} & \textbf{73.7} & \textbf{56.5} & \textbf{94.9} & \textbf{88.7} & \textbf{36.1}   & \textbf{73.8}  & \textbf{71.2} & \textbf{73.0}     \vspace{-0.6mm} \\
    \bottomrule
    \end{tabular}
    }
    \label{tab:multimodal_benchmark}
    \vspace{-0.1em}
\end{table*}

\begin{table*}
    [t!]
    \scriptsize
    \caption{Comparison with existing MLLMs on  long context MLLM benchmarks. ``\#Param'' denotes the number of parameters. The designation ``$-$'' indicates that the corresponding score is not released.}
    
    \vspace{-1.0em}
    \centering
    \renewcommand{\arraystretch}{1.0}
    \resizebox{0.7\linewidth}{!}{
    \begin{tabular}{l c|c c cc c c cc}
    \toprule
    \multirow{2}{*}{Model} & \multirow{2}{*}{\#Param} & \multicolumn{3}{c}{MM-NIAH} & \multicolumn{4}{c}{Milebench} &                         \multirow{2}{*}{VideoMME} \\
    
    \cmidrule(lr){3-5} \cmidrule(lr){6-9}
    &                        & Image       & Text         & Avg          & T             & S             & NI               & Avg & \\
    \midrule
    InternVL2-2B~\cite{internvl_1_5}  & 2.0B       & 23.0        & 18.9         & 21.0         & 58.2          & 54.5          & 37.0             & 49.9                  & $-$  \\
    Phi-3-Vision~\cite{phi}  & 2.7B     & $-$         & $-$          & $-$          & 46.9          & 50.0          & $-$              & $-$                   & $-$   \\
    OmChat~\cite{Omchat}        & 3.9B     & $-$         & $-$          & $-$          & 51.4          & 52.0          & $-$              & $-$                   & 45.9  \\
    LongLLaVA~\cite{LongLLaVA}     & 9B       & $-$         & $-$          & $-$          & 47.3          & 46.8          & $-$              & $-$                   & 43.7   \\
    LongLLaVA~\cite{LongLLaVA}     & 13B      & $-$         & $-$          & $-$          & 52.7          & 52.1          & $-$              & $-$                   & 51.6 \\
    VILA~\cite{Vila}          & 13B      & 14.5        & 40.5         & 27.5         & $-$           & $-$           & $-$              & $-$                   & $-$  \\ 
    Gemini-1.5~\cite{gemini_1_5}    & $-$      & 28.5        & 82.1         & 55.2         & 50.2          & 58.3          & 97.9             & \textbf{68.8}                  & \textbf{69.6}  \\
    GPT-4V~\cite{gpt4v}        & $-$      & $-$         & \textbf{84.1}         & $-$          & 45.6          & 58.9          & \textbf{99.4}             & 68.0                  & 59.9  \\ 
    GPT-4o~\cite{gpt4o_web}        & $-$      & $-$         & $-$          & $-$          & 56.2          & \textbf{63.5}          & $-$              & $-$                   & 64.7   \\ 
    Claude3-Opus~\cite{claude3series}  & $-$      & $-$         & $-$          & $-$          & 37.4          & 48.1          & 85.3             & 56.9                  & 59.7    \\ 
    \midrule
    \rowcolor{gray!15}
    \textbf{InternVL2-V2PE-32K}   & 2.0B & \textbf{72.3}        & 68.1         & \textbf{70.2}         & \textbf{58.5}          & 58.2          & 51.1             & 55.9                   & 48.2     \\
    \bottomrule
    \end{tabular}
    }
    \label{tab:long_benchmark}
    \vspace{-0.2em}
\end{table*}

\subsection{Effectiveness of Our Proposed V2PE}

\textbf{Experimental Setup.}  
We fine-tune the InternVL2-2B model using our proposed V2PE method in a two-stage training procedure. 
1) In the first stage, we fine-tune the released InternVL2-2B model on our mixed training dataset, incorporating V2PE. During training, as described in Eq.~\ref{delta_set}, the positional increment 
\( \delta \) is dynamically selected, enabling the model to flexibly adapt to varying positional increments for visual tokens. The model obtained after this stage is denoted as \textit{InternVL2-V2PE-32K}.
2) In the second stage, we fine-tune the model on the Long-MR-256K dataset, whose sequence lengths  up to 256K token. To preserve the model's general performance, we retain 50\% of the data from the first stage. To optimize memory usage, we apply the Ring Attention~\cite{ringattn}, enabling model parallelism across multiple GPUs. The model obtained after the second stage is denoted as \textit{InternVL2-V2PE-256K}.

\vspace{0.2em}
\noindent\textbf{Evaluation Benchmarks.}  
We evaluate our trained models on various long-context and standard multimodal benchmarks, including the Long-VQA validation split, MM-NIAH~\cite{mmniah} (including our augmented MM-NIAH\textsubscript{1M}), and several widely adopted VQA benchmarks such as ChartQA~\cite{ChartQA}, DocVQA~\cite{DocVQA}, AI2D~\cite{ai2d}, InfoQA~\cite{ai2d}, SQA~\cite{ScienceQA}, POPE~\cite{PoPE}, MMMU~\cite{mmmu}, MMBench\textsubscript{EN}~\cite{mmbench}, and SEED\textsubscript{Image}~\cite{Seed-bench}. 
To systematically assess the VLMs' performance on long-context tasks, we also compare our model with other state-of-the-art methods on two additional long-context benchmarks: MileBench~\cite{milebench} for multi-image and video understanding, and Video-MME~\cite{VideoMME} for video analysis tasks.
For detailed benchmark descriptions and their evaluation metrics, please refer to the appendix.

\vspace{0.2em}
\noindent \textbf{Effectiveness of V2PE.}  
We compare the performance of InternVL2-V2PE-32K and InternVL2-V2PE-256K on the Long-VQA validation split, MM-NIAH (including MM-NIAH\textsubscript{1M}), and standard multimodal benchmarks.  The results are shown in Tab.~\ref{tab:exp_longvqa}, FIg.~\ref{tab:v33_detail}, and Fig.~\ref{tab:exp_mmniah}.

We demonstrate how varying positional increments impact the models' performance on these benchmarks. Specifically, MM-NIAH is evaluated on the image-retrieval task where both the ``needle" and the ``question" are images, posing a stringent challenge for multimodal models to handle visual tokens in long-context sequences.

On standard short-context benchmarks (see Fig.~\ref{tab:v33_detail}), the advantage of V2PE is not immediately evident. In fact, we observe a performance drop on certain tasks, such as DocVQA and InfoQA, when the positional increment is reduced. We hypothesize that this is because InternVL2 has adapted to the original position encoding settings during its alignment and supervised fine-tuning (SFT) training.

However, on long-context benchmarks like Long-VQA and MM-NIAH (see Fig.~\ref{tab:exp_longvqa} and Fig.~\ref{tab:exp_mmniah}), V2PE demonstrates a clear advantage. For instance, on Long-VQA, the best performance is typically achieved when the positional increments are reduced to around 1/8 or 1/16. Moreover, as the token sequence length increases, smaller positional increments tend to be optimal.

In the MM-NIAH benchmark, involving longer token sequences,  V2PE  demonstrates optimal performance with positional increments of 1/256. Notably,  InternVL2-V2PE-32K  markedly enhances performance on sequences of 256K tokens, boosting scores from 1.5 to 56.9. This enhancement is also evident in  InternVL2-V2PE-256K, which records the highest scores across various token sequence lengths when using a positional increment of 1/256. For sequences of 512K tokens, the score surges from 10.8 to 95.1. Additionally, with the same positional increment of 1/256, the InternVL2-V2PE-256K model attains a significant score of 64.5 on sequences reaching 1M tokens.

\begin{figure}[t]
    \centering
    \includegraphics[width=0.9\linewidth]{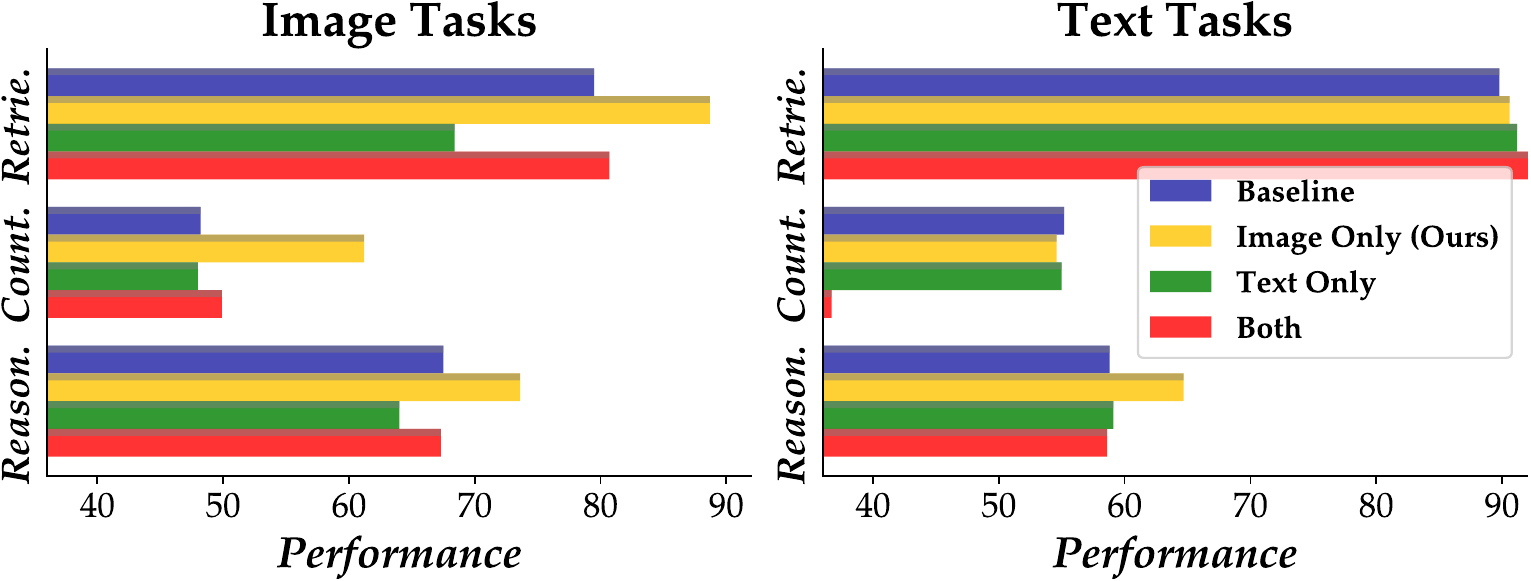}
    \vspace{-0.5em}
    \caption{Performance on MM-NIAH benchmarks with different position encoding strategies. ``Image only", ``Text only", ``Both", and ``Baseline" represent applying V2PE to visual tokens, textual tokens, both simultaneously, and neither, respectively. }
        \label{fig:text}
\end{figure}

\begin{figure}[t]
    \centering
    \includegraphics[width=0.65\linewidth]{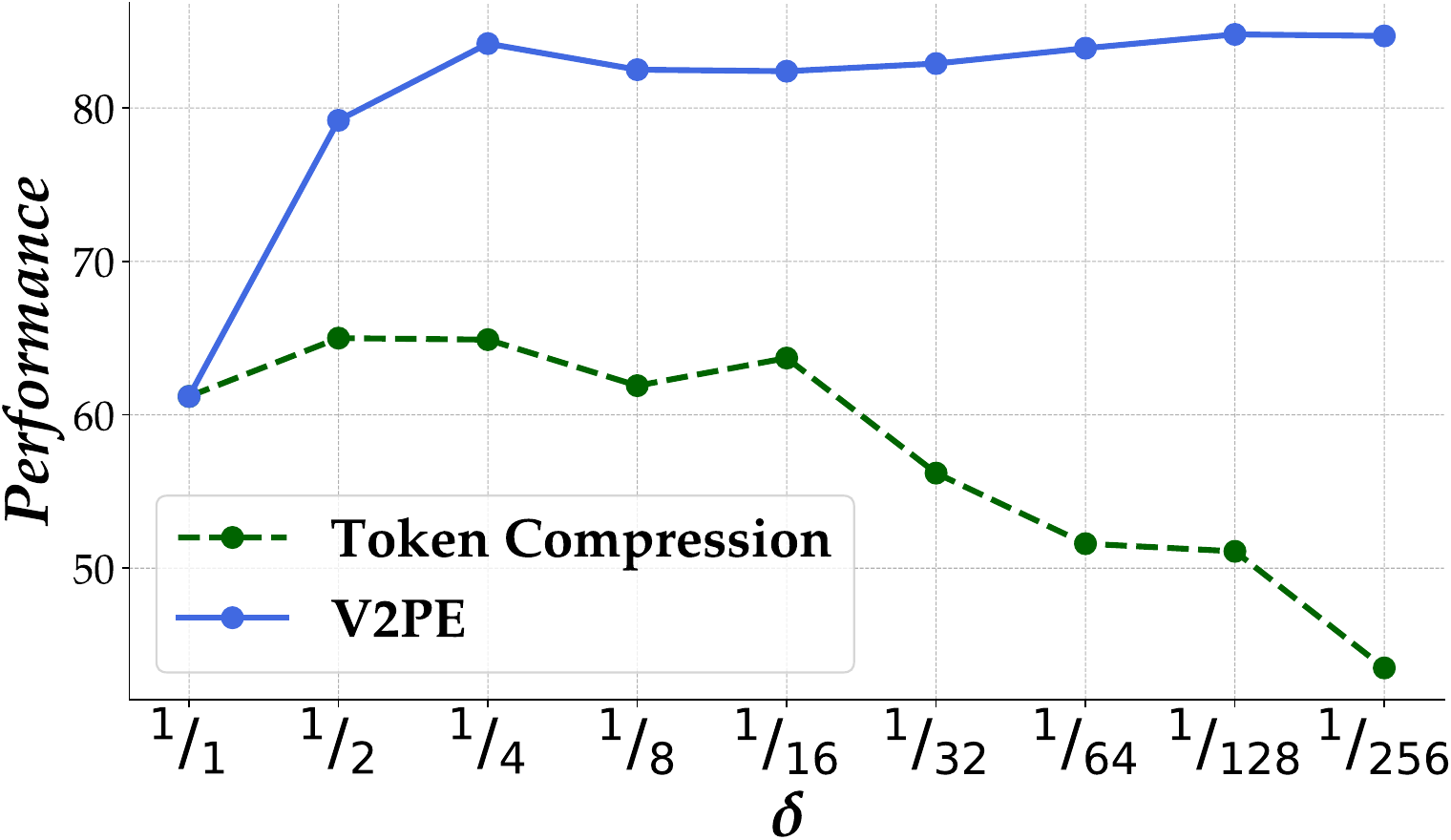}
    \vspace{-0.5em}
    \caption{Performance on image retrieval task in MM-NIAH benchmark using V2PE and token compression strategies. Token compression reduces the number of visual tokens by a ratio $\delta$, whereas V2PE employs a smaller positional increment $\delta$ while preserving all visual tokens.}
    \label{fig:compress_token}
    
\end{figure}

\subsection{Comparison with Other VLMs.}  
We evaluate the performance of InternVL2-V2PE-32K against state-of-the-art vision-language models (VLMs) on both standard and long-context multimodal benchmarks, with results presented in Tab.~\ref{tab:multimodal_benchmark} and Tab.~\ref{tab:long_benchmark}, respectively.
Despite being based on a relatively small 2B parameter model and incorporating substantial long-context training data, InternVL2-V2PE-32K achieves highly competitive results on standard short-context multimodal benchmarks.

On long-context multimodal benchmarks, the model performs exceptionally well, demonstrating that V2PE can significantly enhance long-context processing abilities even with a smaller model. These results validate the potential of V2PE for improving the performance of VLMs on a wide range of long-context multimodal tasks.

\subsection{Ablation Study}

\noindent \textbf{Can V2PE be applied to textual tokens?}  
We apply V2PE  to visual tokens,  to textual tokens, concurrently to both, or to neither, whose results  on six tasks on MM-NIAH  are presented in Fig.~\ref{fig:text}, showing that applying V2PE to textual tokens alone improves the model's language understanding but leads to decreased performance on image-focused tasks. When V2PE was applied to both visual and textual tokens, the model's performance became unstable. However, when V2PE was applied only to visual tokens, the VLM exhibited consistently improves performance across all tasks.

\begin{figure}[t]
    \centering
    \includegraphics[width=0.65\linewidth]{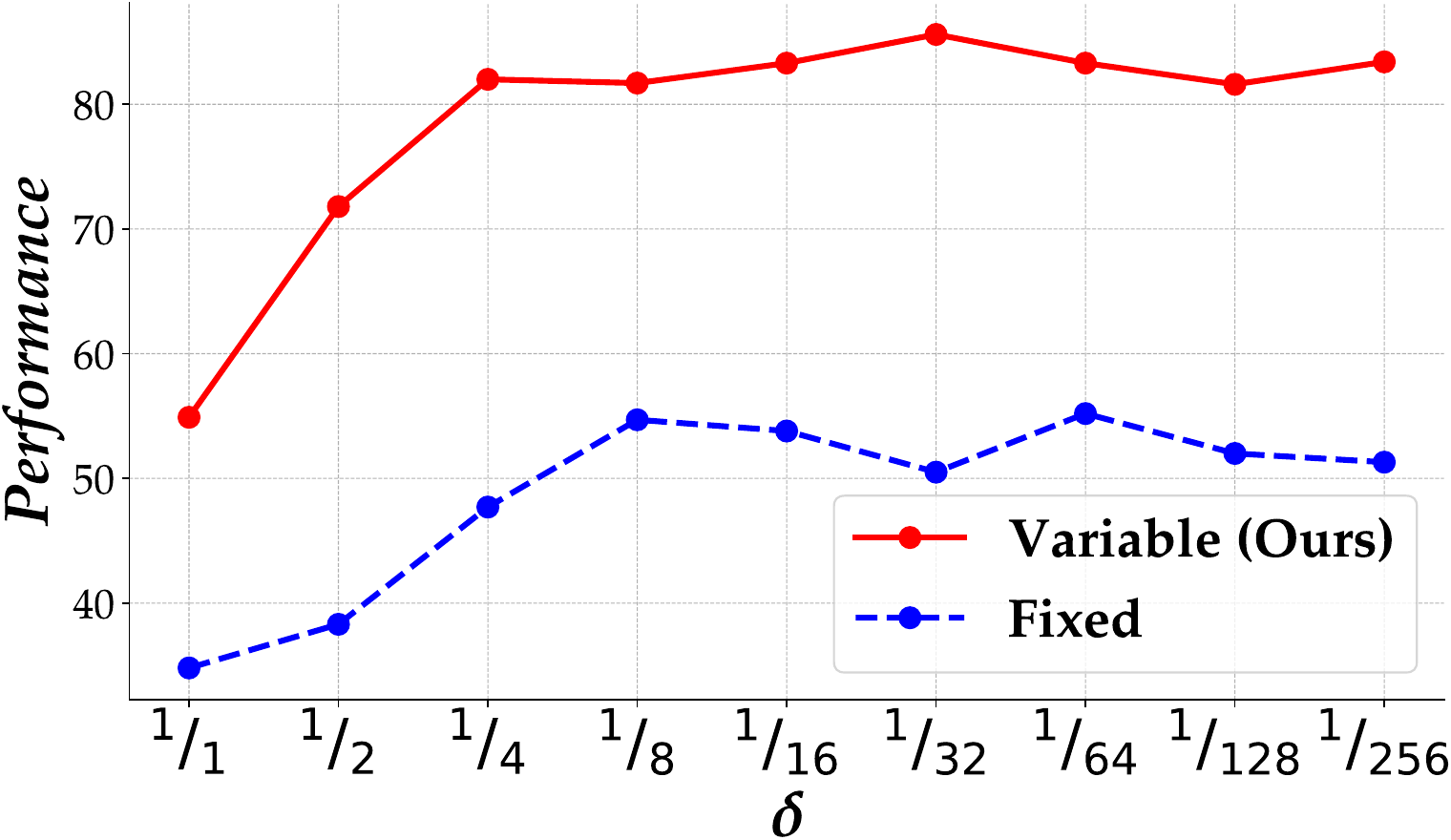}
    \vspace{-0.5em}
    \caption{Performance on image retrieval task in MM-NIAH benchmark with  variable versus fixed position increments for visual tokens in the VLMs.  ``Variable'' denotes training with variable increments, allowing evaluation with arbitrary  $\delta$, while ``Fixed'' means training and evaluating with the same fixed $\delta$.}
    \label{fig:fixed}
    
\end{figure}

\noindent \textbf{Is V2PE equivalent to visual token compression?} 
We also compare V2PE with the token compression strategy, where visual tokens undergo pooling operations to reduce token number, allowing more images to be processed ideally. We tests various compression ratios and evaluate the model's performance on the MM-NIAH image retrieval task. The results, shown in Fig.~\ref{fig:compress_token}, reveal that while token compression results in a rapid performance drop when the number of visual tokens is reduced below 16, our method demonstrates stable performance across all settings. 

\vspace{0.3em}
\noindent \textbf{Can positional increment be fixed during training?}  
We train a set of InternVL2-FT-32K models with fixed positional increments during training. Their performance on the image retrieval task of the MM-NIAH benchmark are shown in Fig.~\ref{fig:fixed}. The results indicate that fixing the positional increment does not lead to optimal task performance. In contrast, the variable adjustment of positional increments provide consistent improvements. 

\vspace{0.3em}

\noindent \textbf{Is V2PE superior to the position encoding extension methods?}  
To compare V2PE with existing position encoding extension methods, we evaluated both InternVL2-FT-32K and InternVL2-V2PE-32K models using two widely adopted position extension techniques: linear interpolation and NTK-Aware Scaled RoPE. The results, presented in Fig.~\ref{fig:stride_line},  Tab.~\ref{tab:exp_longvqa}, and Tab.~\ref{tab:exp_mmniah}, show that V2PE not only offers greater stability but also achieves superior task performance, especially in long-context scenarios.

\section{Conclusion}
\label{sec:conclusion}
We investigate the long-context capabilities of the existing VLM, InternVL2-2B, using our augmented long-context datasets, and find that positional encoding for visual tokens is critical for the long-context capabilities of VLMs. Based on this observation, we introduce V2PE, a novel position encoding strategy that applies smaller, variable positional increments to visual tokens, enabling more efficient handling of long-context multimodal sequences. By leveraging V2PE and our augmented long-context datasets, we fine-tune the open-source InternVL2-2B model successfully, which shows significant improvements on both general and long-context multimodal benchmarks. As for limitations, our approach has not yet been validated on more VLMs or scaled to models with more parameters, due to limited computational resources.

\noindent  \textbf{Acknowledgement}
The work is supported by the National Key R\&D Program of China (NO. 2022ZD0161300), by the National Natural Science Foundation of China (62376134).

{\small
\bibliographystyle{plain}
\bibliography{egbib}
}

\appendix
\section*{Appendix}

\section{Dataset Details} \label{app}

\begin{figure*}[htbp]
    \centering
    \includegraphics[width=0.8\linewidth]{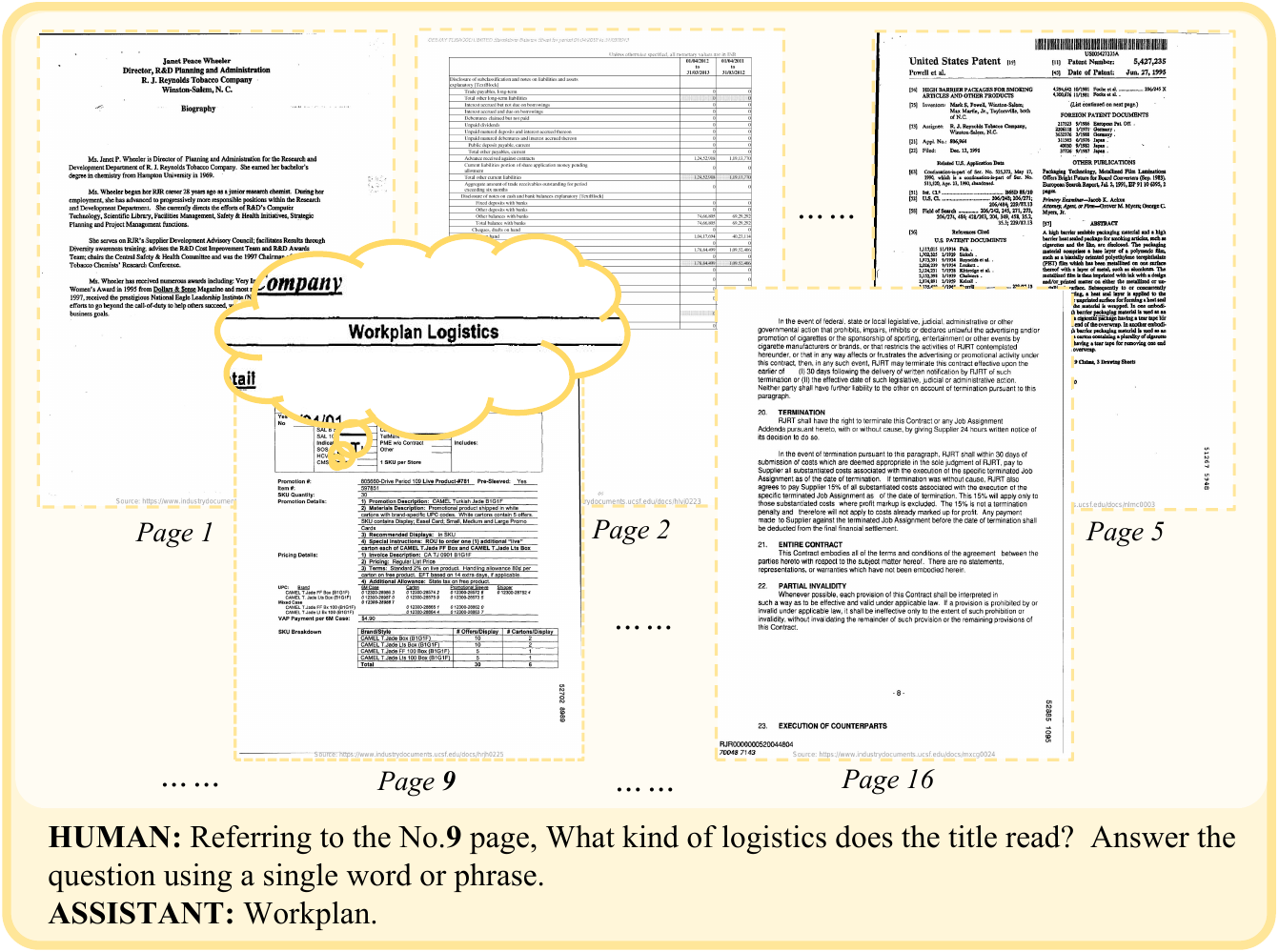}
\caption{Examples of DocVQA subset from Long-VQA dataset.}
    \label{fig:long_docvqa_ex}
\end{figure*}

\begin{figure*}[htbp]
    \centering
    \includegraphics[width=0.8\linewidth]{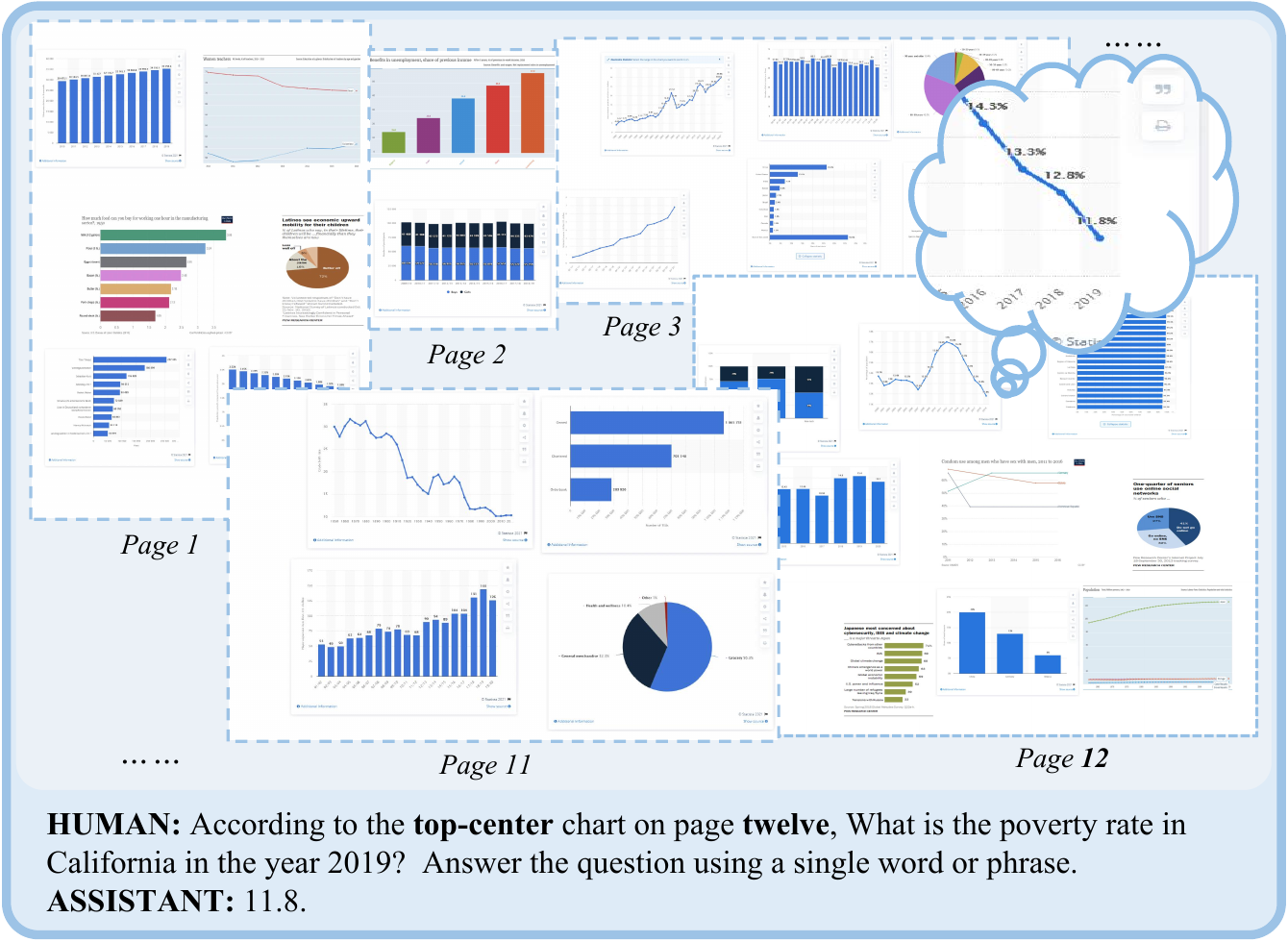}
\caption{Examples of ChartVQA subset from Long-VQA dataset.}
    \label{fig:long_chartvqa_ex}
\end{figure*}

\begin{figure*}[htbp]
    \centering
    \includegraphics[width=0.72\linewidth]{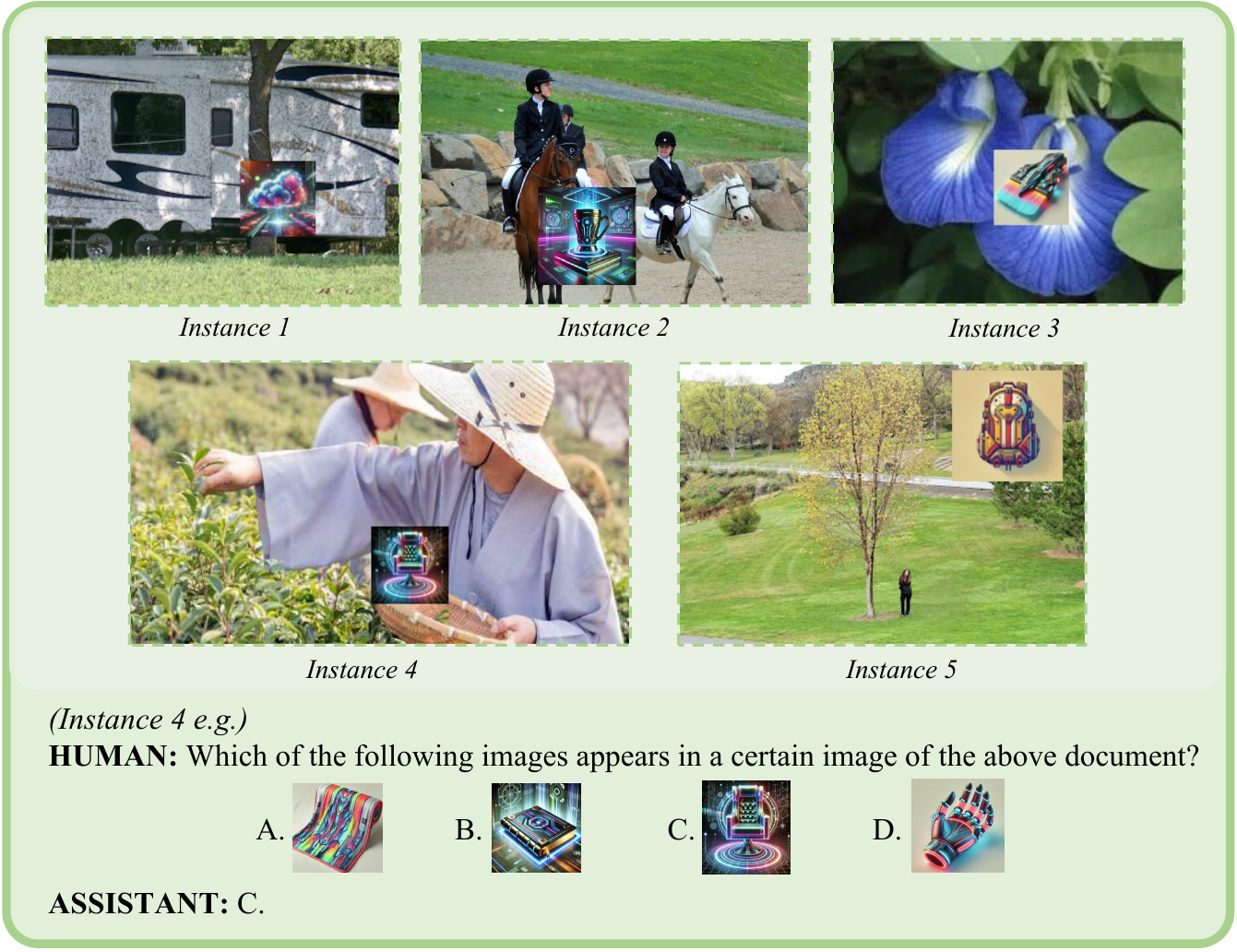}
\caption{Examples of \textit{Retrieval-Image-Needle} in our proposed Long-MR dataset.}
    \label{fig:niah_fake}
\end{figure*}

\begin{figure*}[htbp]
    \centering
    \includegraphics[width=0.72\linewidth]{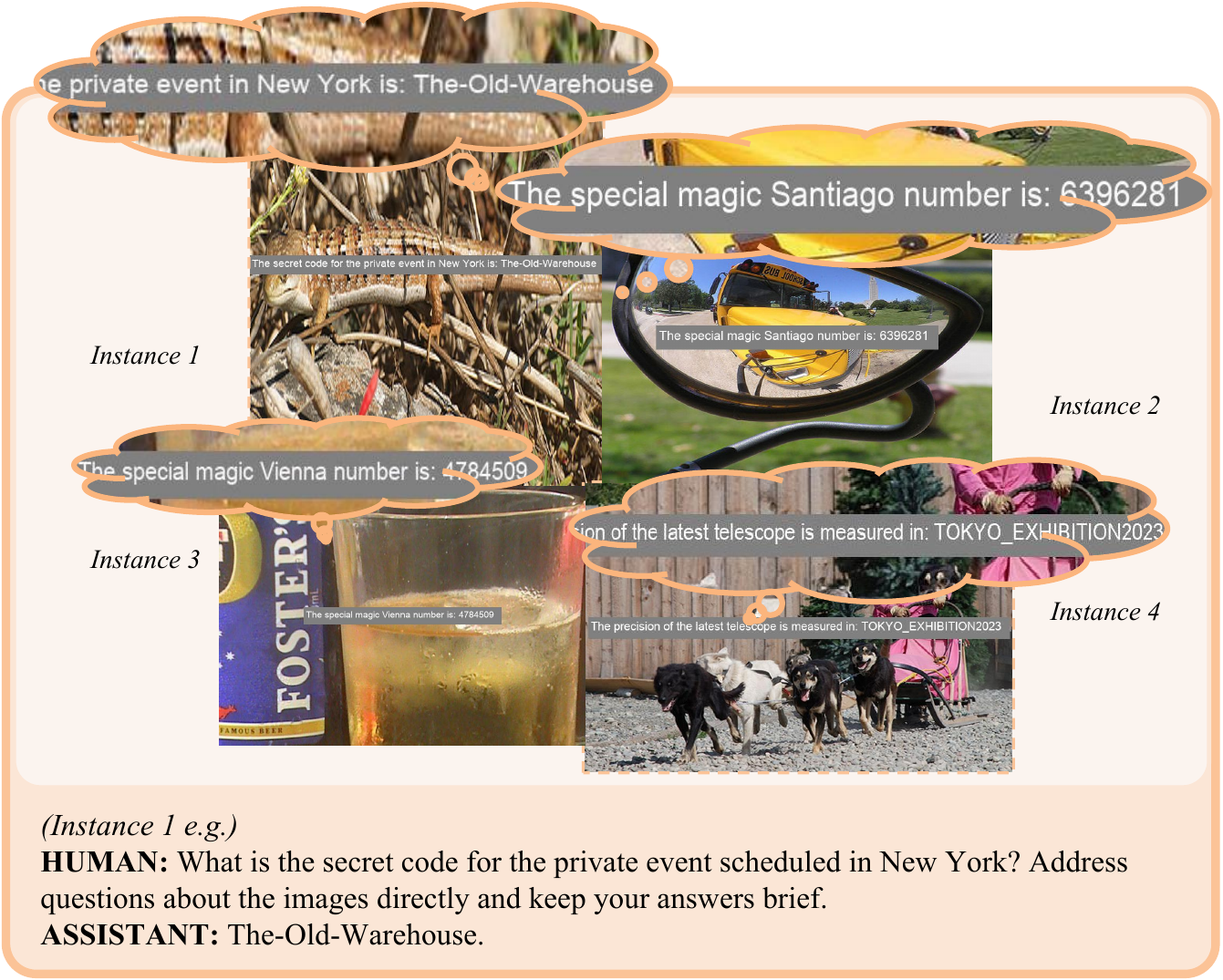}
\caption{Examples of \textit{Image-Needle-In-A-Haystack} with complex needles in our proposed Long-MR dataset. These needles vary in answer format, font-size and style.}
    \label{fig:niah_hard}
\end{figure*}

We have introduced two  augmented long-context multimodal datasets: Long-VQA and Long-MR, designed to systematically evaluate and analyze the long-context capabilities  of Vision-Language Models (VLMs). Representative examples from these datasets are illustrated in Fig.\ref{fig:long_docvqa_ex}, Fig.\ref{fig:long_chartvqa_ex}, Fig.\ref{fig:niah_fake}, and Fig.\ref{fig:niah_hard}. 
Next, we will provide a detailed description of the dataset construction process.

\subsection{Long Visual Question Answering (Long-VQA)}

The Long-VQA dataset presents a novel challenge to VLMs, necessitating advanced visual perception  and sophisticated  reasoning capabilities to address tasks involving long context. 
This dataset is synthesized by combining multiple existing datasets in Tab.~\ref{tab:longdoc_size} to create a set of complex multi-image tasks. 

For datasets that primarily consist of document-like images, such as DocVQA~\cite{DocVQA}, we extend the context by merging multiple single-page documents into cohesive multi-page collections. Questions are subsequently sampled from one of the original documents, ensuring that the model's ability to retain and utilize information across an extended multi-page context is rigorously evaluated.

In the case of datasets composed of visual elements like images, charts, and tables, such as those from GQA~\cite{GQA}, VizWiz~\cite{gurari2018vizwiz}, and TabFact~\cite{Tabfact}, we aggregate these components into complex, multi-page documents that emulate naturalistic scenarios. Each visual element, whether an image or a chart, is strategically positioned across different pages and at various locations (e.g., upper-left, center, lower-right). This configuration is designed to evaluate a model's complex reasoning capabilities, as it requires an understanding of the relative positioning of elements throughout the entire document.

\begin{figure}[htbp]
    \centering
    \includegraphics[width=0.8\linewidth]{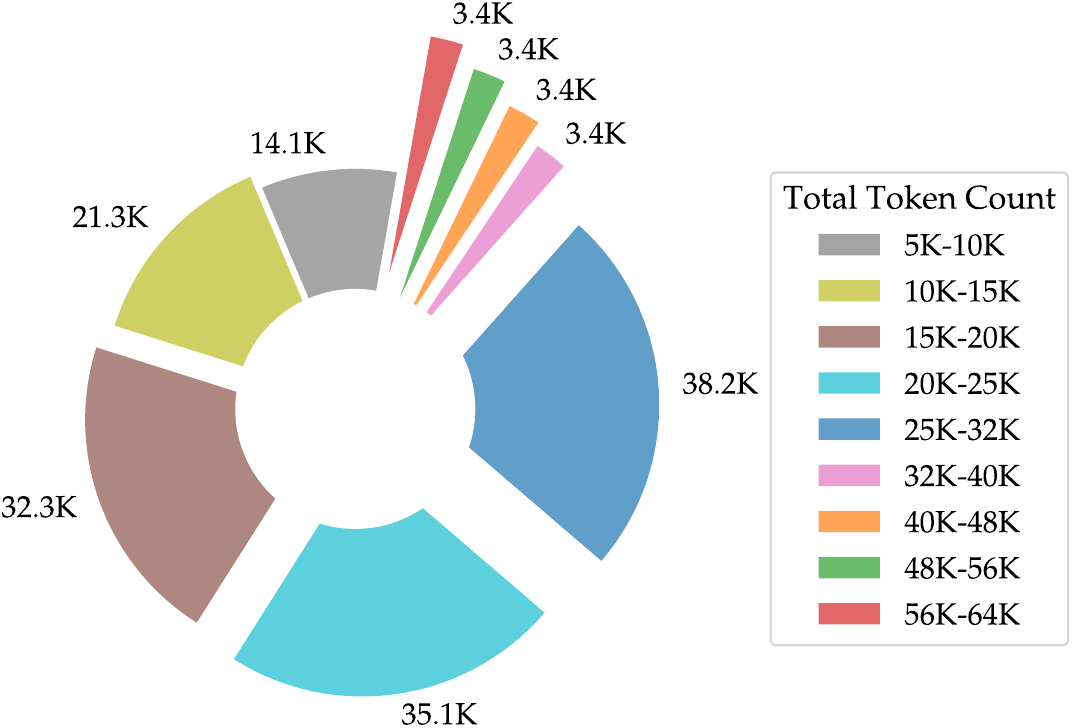}
\caption{The token length distribution of test set in Long-VQA.}
    \label{fig:plot_test_pdf}
\end{figure}

By constructing a diverse and challenging dataset, Long-VQA not only evaluates a model's ability to process a wide range of visual inputs but also emphasizes the necessity of navigating through complex, multi-image contexts. This synthesis of data from multiple sources, combined with the deliberate complexity of the spatial layouts, establishes a rigorous benchmark for VLMs.
Additionally, we provide the length distribution of the Long-VQA test set in Fig.~\ref{fig:plot_test_pdf}.

\subsection{Long Multimodal Retrieval (Long-MR)}

Our proposed Long-MR dataset is constructed upon the MM-NIAH~\cite{mmniah} benchmark, designed specifically to evaluate the performance of VLMs in long-context multimodal retrieval tasks. 
To further assess the generalization capabilities of VLMs within this task, we introduce additional synthetic variations that increase the task complexity.

Unlike the original MM-NIAH, where a single needle is inserted, our Long-MR dataset incorporates multiple needles into the long-context multimodal sequence. 
Of these needles, only one is considered as the target query, while the remainder serve as negative needles. This configuration introduces significantly more challenging negative instances, compelling the model to accurately distinguish between highly similar yet irrelevant needles in a lengthy contextual sequence.

To enrich the diversity of needles, we leverage advanced large language models (LLMs) to create synthetic needles beyond those included in the official MM-NIAH benchmark. This expansion results in a more heterogeneous dataset that emulates real-world complexity, thereby improving the robustness of the evaluation. Such diversification reduces the risk of the model overfitting to a particular needle category, fostering the development of more generalizable retrieval capabilities.

\begin{figure}[htbp]
    \centering
    \includegraphics[width=0.8\linewidth]{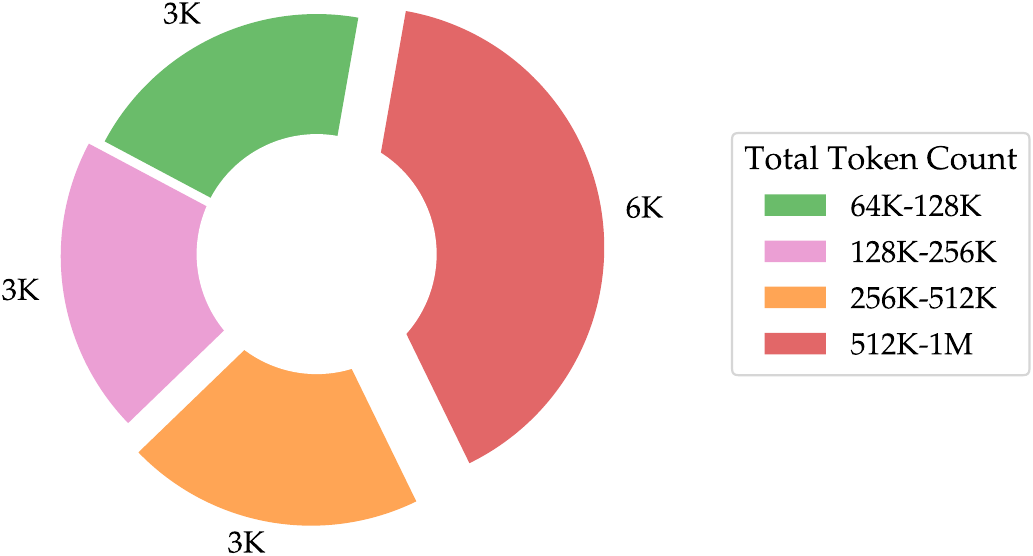}
\caption{The token length distribution of our MM-NIAH\textsubscript{1M}}

    \label{fig:plot_test_niah_img}
\end{figure}

Fig.~\ref{fig:plot_test_niah_img} illustrates the length distribution of our costumed evaluation split, denoted as MM-NIAH\textsubscript{1M}. Notably, the majority of sequences fall within the 512K to 1M token range. For contexts with lengths shorter than 64K, we directly utilize the samples from the original MM-NIAH benchmark.

\begin{table}
\caption{Data statistics of Long-VQA dataset.}
    \centering
    \small
\begin{tabular}{lcc}
\toprule
Dataset &  \multicolumn{2}{c}{Dataset Size}   \\
\midrule
 & Training & Validation \\
\hline 
\centering
DeepForm~\cite{deepform} & 3.4K & 2.1K \\
DocVQA~\cite{DocVQA} & 39.4K & 6.0K \\
InfoVQA~\cite{infoqa} & 23.9K & 4.1K \\
Kleister~\cite{Kleister} & 13.4K & 5.5K \\
SQA~\cite{ScienceQA} & 10.2K & 4.1K \\
VisualMRC~\cite{Visualmrc} & 15.8K & 7.4K\\
ChartQA~\cite{ChartQA} & 40.1K & 3.3K\\
DVQA~\cite{Dvqa} & 150.0K & 16.4K \\
TabFact~\cite{Tabfact} & 91.6K & 13.4K \\
WikitabQS~\cite{wikitabQS} & 14.1K & 5.1K \\
Clevr~\cite{Clevr} & 150.0K & 16.4K \\
GQA~\cite{GQA} & 150.0K & 16.4K \\
OcrVQA~\cite{Ocr-vqa} & 150.0K & 16.4K \\
OKVQA~\cite{okvqa} & 9.0K & 5.8K \\
TextCaps~\cite{Textcaps} & 110.0K & 17.2K \\
TextVQA~\cite{textvqa} & 56.5K & 6.5K \\
Vizwiz~\cite{gurari2018vizwiz} & 20.5K & 8.8K \\
\hline Total & 1.1M & 155.0K \\
\bottomrule
\end{tabular}
    \label{tab:longdoc_size}
\end{table}

\section{Evaluation} \label{sup:eval}

To address the out-of-memory challenge encountered during inference on samples exceeding token lengths of 128K in the MM-NIAH\textsubscript{1M} evaluation dataset, we adopt a perplexity-based approach similar to that employed in LongVA~\cite{LongVA}. Specifically, during evaluation, we concatenate the question embedding, which integrates both textual and visual components, with the output answer embedding. Subsequently, a single forward pass is performed using ring attention to predict the logits of the answer. The output is considered correct if the index corresponding to the highest output logit across all tokens within the answer span aligns with the correct answer.

To facilitate comparison between  position encoding extension  and V2PE, we determine the interpolation factor for linear interpolation~\cite{PI} based on the test sample length and the context window size used during training. Specifically, we interpolate the position indices of the test samples to match the context window range from the training phase. For example, when evaluating InternVL2-FT-32K on the 64K-length Long-VQA task using linear interpolation~\cite{PI}, we utilize an interpolation factor of 2, which effectively maps the position indices of the test samples into the 32K range, consistent with the context length employed during training. Similarly, for evaluations involving 1M-length samples, an interpolation factor of 32 is selected.
For the NTK-Aware Scaled RoPE~\cite{ntk}, we fix the scaling factor at 5, as our experimental results indicate that it yields consistent performance across tasks of varying lengths.

\begin{table}
\caption{Summary of our training hyper-parameters.}
    \centering
    \small
\begin{tabular}{ll}
\toprule
Configuration &  V2PE Setting   \\
\midrule
\centering
Weight init & InternVL2-2B~\cite{internvl_1_5} \\
Loss type & Generative loss \\
Learning rate schedule & Cosine decay \\
Optimizer & AdamW~\cite{adamw} \\
Learning rate & 5e-6 \\
Weight decay & 5e-2 \\
Input image resolution & 448 $\times$ 448 \\
Warmup  steps & 150\\
Iterations & 5K \\
\bottomrule
\end{tabular}
    \label{tab:exp_setting}
\end{table}

\section{Experiment Details} \label{sup:exp_details}

The detailed training configurations are summarized in Table~\ref{tab:exp_setting}. Additionally, for experiments involving the V2PE method, we employ the \texttt{Float32} data type when computing positional indices and positional embeddings required for RoPE, to ensure computational precision.

\section{Attention Matrices Analysis} \label{sup:attn}

To investigate the impact of our V2PE on attention mechanism, we follow  ~\cite{csordas2021neural} to analyze the attention matrices on the Long-VQA evaluation set. Specifically, our analysis focuses on the tail portion of the entire attention matrices, which corresponds to the question segments located at the end of the sequences. This allows us to observe how effectively the model retrieves relevant information when answering questions.

\begin{figure*}[htbp]
    \centering
    \includegraphics[width=1.0\linewidth]{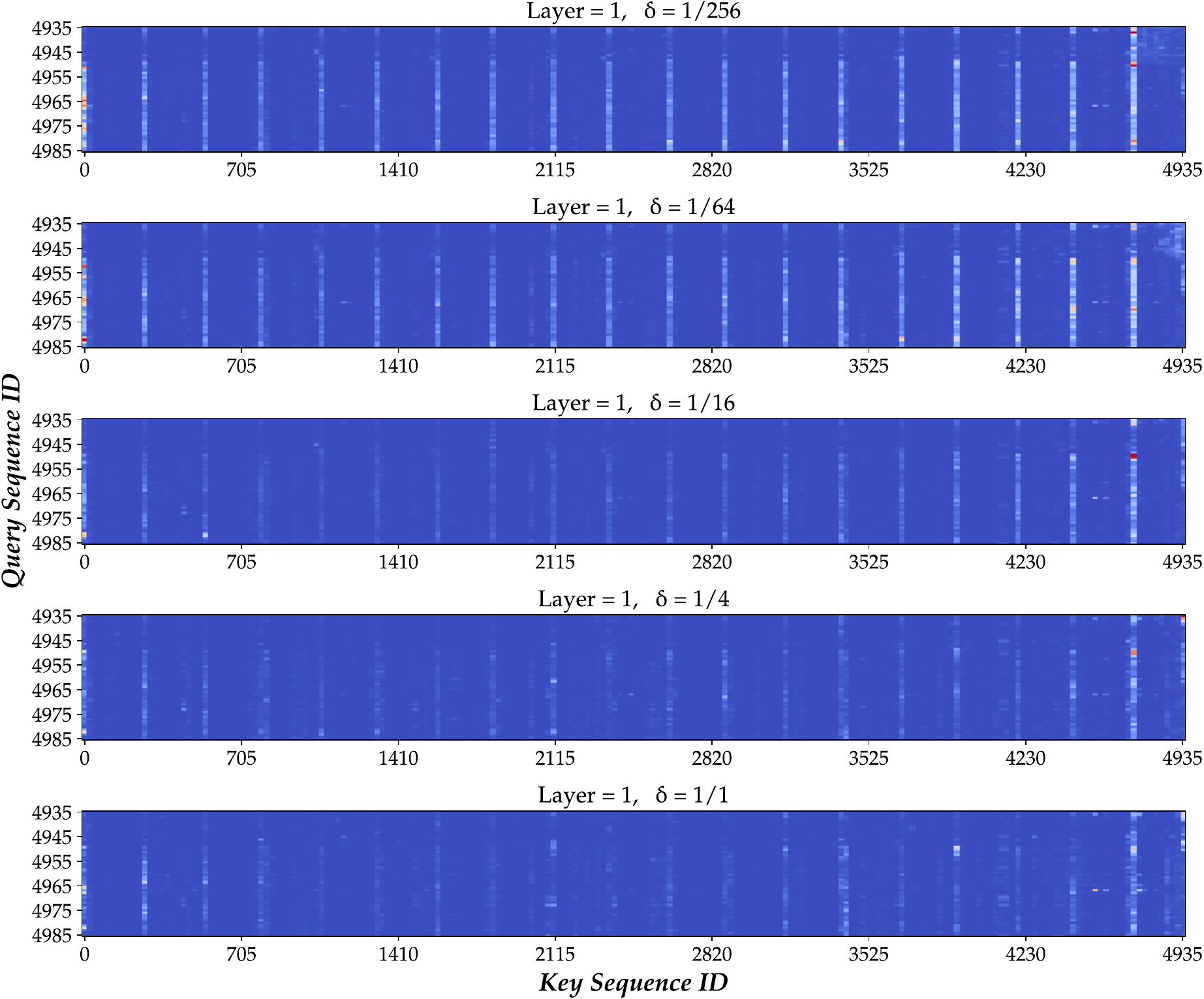}
\caption{Attention map visualization in layer 1 (Maximum over 16 heads).}
    \label{fig:atten_layer_2}
\end{figure*}

\begin{figure*}[htbp]
    \centering
    \includegraphics[width=1.0\linewidth]{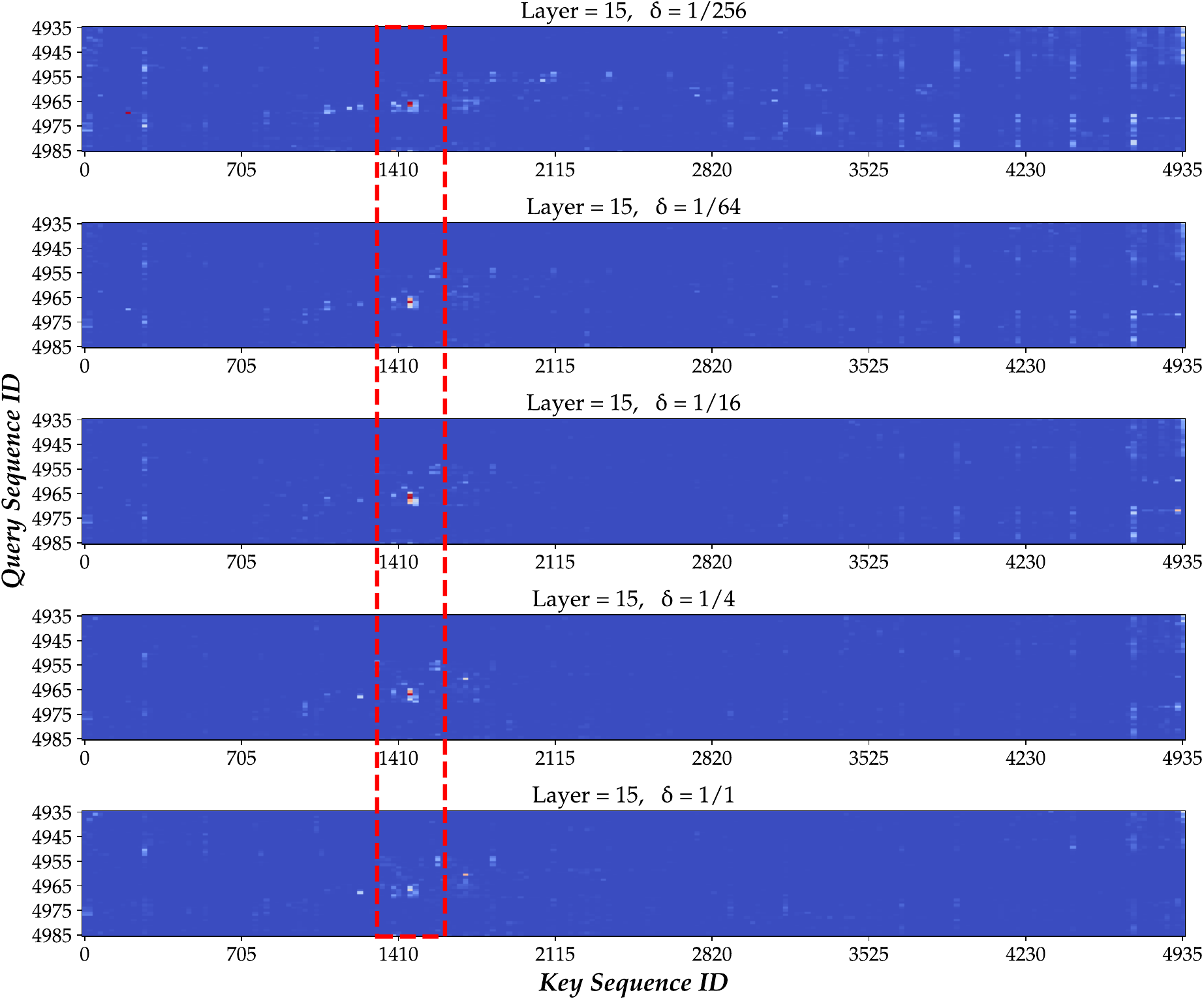}
\caption{Attention map visualization in layer 15 (Maximum over 16 heads).}
    \label{fig:atten_layer_15}
\end{figure*}

As illustrated in Fig.~\ref{fig:atten_layer_2}, we observe that as the positional increment parameter $\delta$ decreases,  the attention patterns in Layer 1 exhibit an increasingly distinct emphasis on visual tokens.
This observation suggests that with smaller values of $\delta$, the model becomes more attentive to visual content, which is crucial for answering questions involving visual inputs.
Furthermore, Fig.~\ref{fig:atten_layer_15} shows that in deeper layers (e.g., Layer 15), the attention becomes more focused around a specific sequence index, particularly ID${=}$1410, as $\delta$ decreases. Notably, the answer to the corresponding question is located near the 1410-\textit{th} token. This indicates that a smaller $\delta$ not only sharpens the model's focus but also aligns its attention more effectively with the tokens containing the correct answer.

These findings imply that using smaller positional increment $\delta$ allows the model to better align its attention with the critical portions of the input sequence, thereby enhancing its capability to retrieve relevant information, especially in the   scenarios of long-context multimodal tasks.

\end{document}